\documentclass[10pt,twocolumn,letterpaper]{article}

\usepackage[table]{xcolor}
\usepackage[pagenumbers]{cvpr} % To force page numbers, e.g. for an arXiv version

\usepackage{multirow}
\usepackage{algorithmic}
\usepackage{algorithm}
\definecolor{cvprblue}{rgb}{0.21,0.49,0.74}
\definecolor{cGreen}{RGB}{100,180,100}
\definecolor{cRed}{RGB}{220,50,0}
\definecolor{Klein_Blue}{rgb}{0.0, 0.129, 0.6}
\definecolor{mygray2}{gray}{0.9}
\definecolor{mygray1}{gray}{0.95}
\definecolor{brown10}{rgb}{0.9, 0.84, 0.80}  % 这个是
\definecolor{brown20}{rgb}{0.95, 0.90, 0.85}
\usepackage[pagebackref,breaklinks,colorlinks,allcolors=cvprblue]{hyperref}

\def\paperID{219} % *** Enter the Paper ID here
\def\confName{CVPR}
\def\confYear{2026}

\title{UETrack: A Unified and Efficient Framework for Single Object Tracking}

\author{
Ben Kang$^1$, Jie Zhao$^{1,*}$, Xin Chen$^2$, Wanting Geng$^1$, 
Bin Zhang$^1$, Lu Zhang$^1$, Dong Wang$^1$, Huchuan Lu$^1$\\
$^1$Dalian University of Technology\\
$^2$City University of Hong Kong\\
{\tt\small \{kangben, gengwanting, binzhang\}@mail.dlut.edu.cn, xche32@cityu.edu.hk}\\
{\tt\small luzhangdut@gmail.com, \{zj982853200, wdice, lhchuan\}@dlut.edu.cn}
}

\begin{document}
\maketitle
\begin{abstract}%
With growing real-world demands, efficient tracking has received increasing attention. However, most existing methods are limited to RGB inputs and struggle in multi-modal scenarios. Moreover, current multi-modal tracking approaches typically use complex designs, making them too heavy and slow for resource-constrained deployment.
To tackle these limitations, we propose \textbf{UETrack}, an efficient framework for single object tracking. 
UETrack demonstrates high practicality and versatility, efficiently handling multiple modalities including RGB, Depth, Thermal, Event, and Language, and addresses the gap in efficient multi-modal tracking.
It introduces two key components: a Token-Pooling-based Mixture-of-Experts mechanism that enhances modeling capacity through feature aggregation and expert specialization, and a Target-aware Adaptive Distillation strategy that selectively performs distillation based on sample characteristics, reducing redundant supervision and improving performance.
Extensive experiments on 12 benchmarks across 3 hardware platforms show that UETrack achieves a superior speed–accuracy trade-off compared to previous methods. 
For instance, UETrack-B achieves 69.2\% AUC on LaSOT and runs at 163/56/60 FPS on GPU/CPU/AGX, demonstrating strong practicality and versatility.
Code is available at \href{https://github.com/kangben258/UETrack}{https://github.com/kangben258/UETrack}.
\end{abstract}
\renewcommand{\thefootnote}{}
\footnote{$^*$Corresponding author.}
\section{Introduction}
Single Object Tracking (SOT) is a fundamental task in computer vision that aims to continuously locate a specified target in a video.
Recently, efficient trackers have gained increasing attention due to their higher practicality compared with mainstream trackers~\cite{aqatrack, PrDiMP, mixformer, TATrack, mixformer_journal, odtrack, LoRAT, artrackv2, evptrack}. However, most existing efficient trackers~\cite{anti-uav, HiT, mixformerv2} are restricted to RGB-only scenarios, and little effort has been devoted to efficient multi-modal tracking.
In complex real-world environments, a single modality is often insufficient. To enhance robustness, additional modalities such as depth, thermal, or event data are needed.
Although several studies~\cite{sutrack, OneTracker, SDSTrack} have explored multi-modal tracking, the heterogeneity among modalities makes it challenging to effectively capture complementary information and shared representations. Consequently, existing methods often depend on complex designs and large model structures, resulting in high computational cost and latency that hinder their deployment in real-world applications.
These limitations raise a key question: \textit{Can we design an efficient multi-modal tracking model suitable for real-world scenarios?}

\begin{figure}[t]
\begin{center}
\includegraphics[width=1\linewidth]{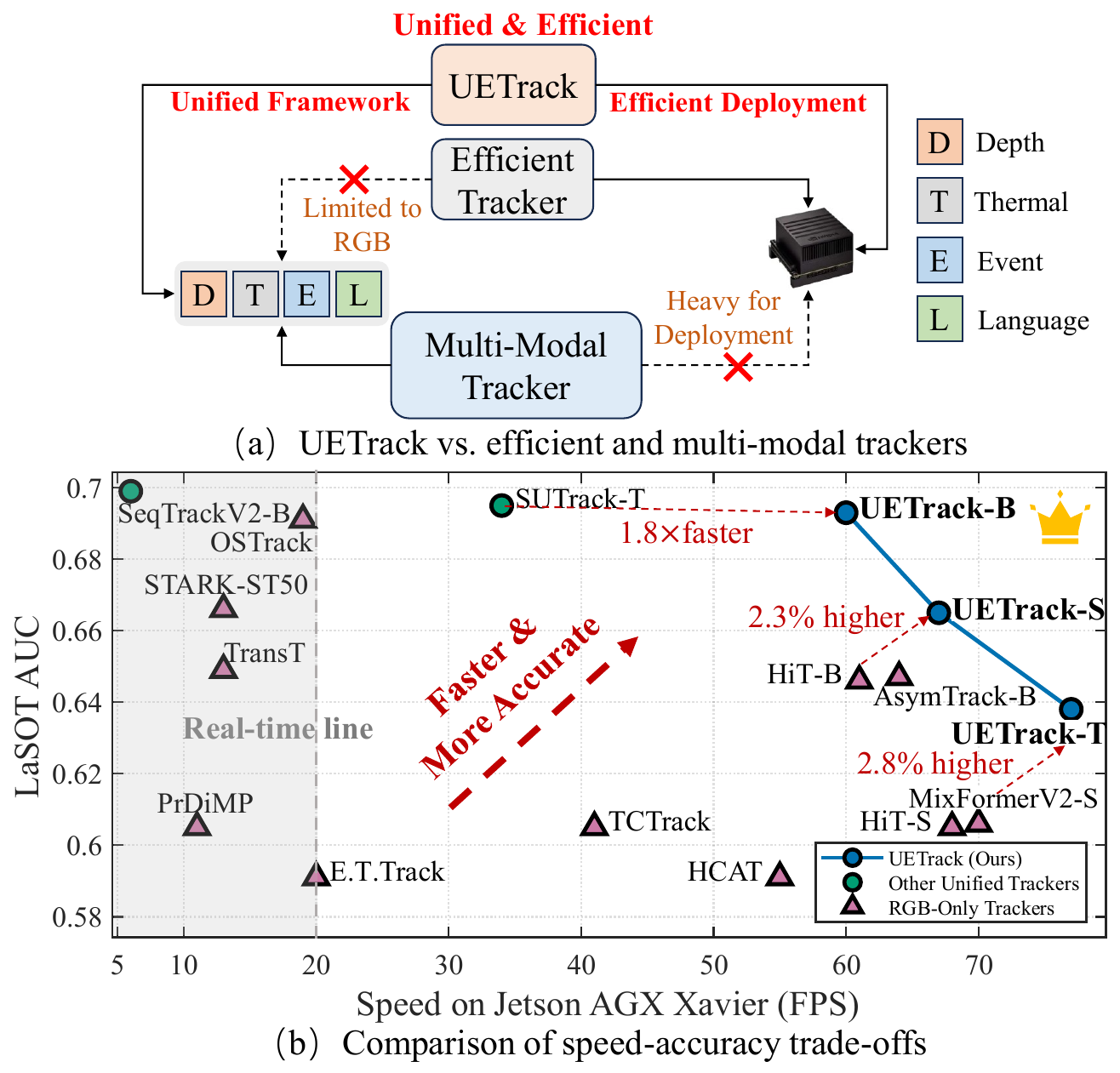}
\end{center}
\vspace{-3mm}
   \caption{UETrack vs. Other Trackers. (a) compares UETrack with current efficient and multi-modal trackers; (b) presents a comparison of speed-accuracy trade-offs on the Jetson AGX.} 

\label{fig:pipeline}
\vspace{-4mm}
\end{figure}

%%%%
To address the above issues, we propose an efficient SOT framework named \textbf{UETrack}. 
As shown in Figure~\ref{fig:pipeline}(a), UETrack adopts a lightweight architecture and supports multiple modalities, including RGB, Depth, Thermal, Event, and Language. This design enables efficient multi-modal tracking with strong practicality and versatility, making UETrack suitable for real-world applications.
We follow SUTrack~\cite{sutrack} to achieve unified modeling across multiple modalities.
Specifically, for depth, thermal, and event data, we concatenate each with the paired RGB image to form a 6-channel composite input, which is fed into a patch embedding layer to generate image token embeddings. For language data, we leverage CLIP~\cite{clip} to obtain language token embeddings. All embeddings are then jointly processed by the transformer blocks. 
This unified processing pipeline significantly reduces the computational cost of multi-modal modeling and enables efficient inference across all modalities.
Due to the heterogeneity among different modalities, efficient trackers with limited parameters often struggle to capture complementary information and shared representations across modalities. To address this issue, we introduce a Token-Pooling-based Mixture-of-Experts (TP-MoE) structure.
Unlike traditional MoE methods~\cite{moe}, TP-MoE eliminates the complex and time-consuming gating mechanism, and instead adopts a soft assignment strategy via weighted feature aggregation. 
This design enables efficient collaboration and specialization among experts, improving feature modeling in multi-modal scenarios while maintaining high model efficiency.
Additionally, we propose a Target-aware Adaptive Distillation (TAD) strategy to further enhance UETrack’s performance. 
TAD adaptively determines whether a sample requires supervision from the teacher model’s target distributions and feature maps, and dynamically adjusts the degree of distillation. This mechanism filters out misleading signals, mitigating the negative impact of unreliable teacher outputs on the student.

Extensive experiments on 12 datasets and 3 platforms show that UETrack achieves a superior speed–accuracy trade-off across multiple tasks. As shown in Figure~\ref{fig:pipeline}(b), UETrack-B runs 1.8$\times$ faster on AGX and 2.4$\times$ faster on CPU than SUTrack-T~\cite{sutrack}, while maintaining comparable accuracy. UETrack-S improves the AUC on LaSOT by 2.3\% and runs 1.1$\times$ faster on AGX compared to HiT-B~\cite{HiT}. Similarly, UETrack-T achieves a 2.8\% AUC gain on LaSOT over MixFormerV2-S~\cite{mixformerv2}, with a 1.1$\times$ speedup on AGX.
Our contributions are summarized as follows:
\begin{itemize}[leftmargin=0.468cm]
\item{
We propose an efficient SOT framework, UETrack, which can efficiently process RGB, Depth, Thermal, Event, and Language modalities. UETrack demonstrates strong practicality and versatility, filling the gap in efficient multi-modal tracking.
}
\item{We introduce the Token-Pooling-based Mixture-of-Experts (TP-MoE) to enhance the representation ability for multi-modal inputs. Additionally, we propose the Target-aware Adaptive Distillation (TAD) strategy to further boost performance.}
% \item{UETrack represents a family of unified and efficient trackers designed for real-world applications. Extensive experiments verify its effectiveness, and it achieves new state-of-the-art performance.}

\end{itemize}

\section{Related Work}
\noindent\textbf{Efficient Object Tracking.}
Unlike mainstream deep trackers~\cite{liuchang2,ToMP,AiATrack,liuchang1} that prioritize accuracy, efficient trackers aim to balance accuracy and inference speed.
Early works~\cite{hcatm,ATOM,lighttrack} are CNN-based~\cite{AlexNet,ResNet} and achieve high speed, but their accuracy lags behind mainstream models.
With the rise of Transformer architectures~\cite{ViT,swintransformer}, several Transformer-based efficient trackers~\cite{fear,ETTrack,HiT,smat,hift,HCAT,litetrack,asymtrack} have emerged, significantly improving tracking accuracy while maintaining fast inference.
However, most efficient trackers are limited to RGB-only scenarios and underperform in complex environments requiring multi-modal cues. In contrast, our proposed UETrack is a unified framework that supports five modalities, offering improved practicality and versatility.

\noindent\textbf{Multi-Modal Object Tracking.}
% In complex scenarios, a single RGB modality is often insufficient to handle diverse visual disturbances. To enhance tracking robustness, multimodal tracking approaches (e.g., CA3DMS, JMMAC, CapsuleTNL) have been proposed. 
% The dominant types of multimodal tracking include Depth~\cite{ca3dms,dal,LTMU}, Thermal~\cite{jmmac,apfnet,cmpp}, Event~\cite{MDNet,SiamBAN}, and Language~\cite{CapsuleTNL,VLTTT,CTRTNL}, which enhance robustness by leveraging auxiliary modalities.
% Recently, unified models like VipT~\cite{vipt}, Un-Track~\cite{untrack}, SDSTrack~\cite{SDSTrack}, and OneTrack~\cite{OneTracker} adapt RGB trackers with modality-specific modules, while SUTrack~\cite{sutrack} uses unified tokens to process multiple modalities without extra modules.
The dominant types of multi-modal tracking include Depth~\cite{ca3dms,dal}, Thermal~\cite{jmmac,apfnet}, Event~\cite{MDNet,SiamBAN}, and Language~\cite{CapsuleTNL,VLTTT}. By leveraging complementary information from auxiliary modalities, these methods significantly improve performance under challenging conditions.
Recently, unified modeling has emerged, aiming to handle multiple modalities within a single architecture. Models like ViPT~\cite{vipt}, Un-Track~\cite{untrack}, SDSTrack~\cite{SDSTrack}, and OneTracker~\cite{OneTracker} adapt existing RGB trackers by incorporating modality-specific modules, while SUTrack~\cite{sutrack} uses unified tokens to process multiple modalities without extra modules.
However, most of them suffer from complex architectures and high computational cost, limiting practical use. In contrast, our UETrack maintains strong performance with significantly faster inference, offering improved practicality and efficiency.

\noindent\textbf{Knowledge Distillation.}
Knowledge distillation is a common approach to improve efficient model performance. Existing methods include soft distribution distillation~\cite{kd,distbert}, guiding the student to mimic the teacher’s output distribution; feature-based distillation~\cite{fitnet,at}, aligning intermediate representations; and relational distillation~\cite{rkd,skd}, modeling inter-sample relationships. 
Recently, adaptive strategies~\cite{spot,dkd} have gained attention for dynamically reducing redundant supervision and enhancing distillation. In this work, we propose a Target-aware Adaptive Distillation strategy tailored for object tracking, improving the specificity and effectiveness of knowledge transfer.

\begin{figure*}[t]
\begin{center}
\includegraphics[width=\linewidth]{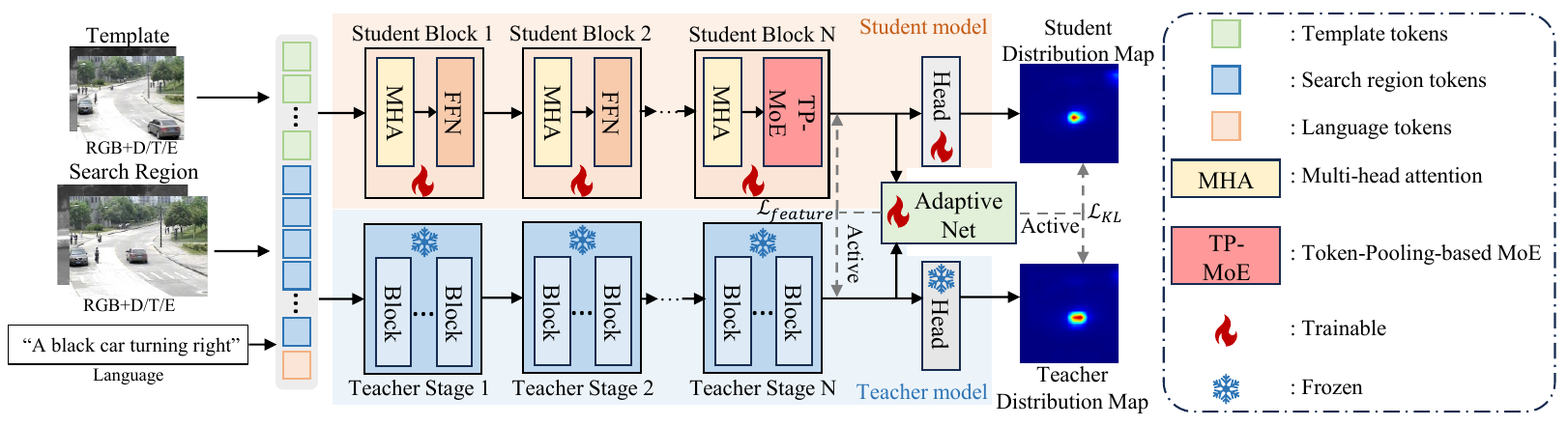}
\end{center}
\vspace{-3mm}
   \caption{Architecture of UETrack. The training pipeline consists of a teacher model, a student model, and an Adaptive Net for adaptive distillation. During inference, only the student model is used, with TP-MoE as the core component to enhance multi-modal modeling.
   } %
\label{fig:framework}
\vspace{-4mm}
\end{figure*}

\noindent\textbf{Mixture of Experts (MoE).}
MoE has emerged as an effective way to expand model capacity while improving computational efficiency, widely adopted in NLP~\cite{switchtransformer,glam}. Recently, MoE extended to vision tasks~\cite{vmoe,softmoe}, where learnable routing integrates into ViT to balance modeling power and efficiency. In tracking, methods like MoETrack~\cite{moetracker}, eMoE-Tracker~\cite{emoe}, and SPMTrack~\cite{spmtrack} leverage MoE to boost performance. However, gating in MOE often introduces latency.
To address this, we propose a Token-Pooling-based MoE that eliminates gating for efficient tracking.

\section{UETrack}

\subsection{Overall Architecture}
The overall architecture of UETrack is illustrated in Figure~\ref{fig:framework}. 
We first build an efficient student model based on Token-Pooling-based Mixture-of-Experts (TP-MoE). To enhance its performance, we further propose a Target-aware Adaptive Distillation (TAD) framework, which uses SUTrack-B~\cite{sutrack} as the teacher model and incorporates an Adaptive Net to enable dynamic supervision. During training, only the student and Adaptive Net are updated, while the teacher remains frozen.

The input to UETrack consists of multiple modalities, including RGB, Depth, Thermal, Event, and Language. To enable efficient multi-modal modeling, we follow the design of SUTrack by encoding different modalities into unified token embeddings, which minimizes parameter redundancy and computational cost.
Specifically, for Depth, Thermal, and Event modalities, the input is formed as an RGB-X image pair, consisting of the original RGB image $\mathbf{I}_{\text{rgb}} \in {\mathbb{R}}^{ H \times W \times 3}$($H$ and $W$ denote the height and width of the image) and an auxiliary modality image $\mathbf{I}_{\text{aux}} \in {\mathbb{R}}^{ H \times W \times 3}$. These two images are concatenated along the channel dimension to create a composite image $\mathbf{I}_{\text{c}} \in {\mathbb{R}}^{ H \times W \times 6}$.
For RGB and Language modalities, which lack corresponding auxiliary images, we replicate the RGB image along the channel dimension to construct $\mathbf{I}_{\text{c}}$. 
The template and search images form $\mathbf{I}_{\text{c}}^z \in {\mathbb{R}}^{ H_{z} \times W_{z} \times 6}$ and $\mathbf{I}_{\text{c}}^x \in {\mathbb{R}}^{ H_{x} \times W_{x} \times 6}$, respectively. These are passed through a patch embedding layer to produce token embeddings $\mathbf{T}_{\text{c}}^z \in {\mathbb{R}}^{D \times {\frac{H{z}}{16}} \times {\frac{W_{z}}{16}}}$ and $\mathbf{T}_{\text{c}}^x \in {\mathbb{R}}^{D \times {\frac{H{x}}{16}} \times {\frac{W_{x}}{16}}}$. The patch embedding process includes a convolutional downsampling layer with stride 4, followed by MLP layers and two convolutional merging layers to construct high-quality token representations. For the language modality, textual information is extracted using a pre-trained CLIP text encoder~\cite{clip}, which outputs language token embeddings $\mathbf{T}_{\text{l}}$. These embeddings are projected to match the image token dimension via a linear transformation. The CLIP encoder remains frozen during training.
Finally, the token embeddings $\mathbf{T}_{\text{c}}^z$, $\mathbf{T}_{\text{c}}^x$, and $\mathbf{T}_{\text{l}}$ are concatenated to form the input sequence $\mathbf{T} \in {\mathbb{R}}^{ L \times D}$ ($L={\frac{H{z}}{16}} \times {\frac{W_{z}}{16}} + {\frac{H_{x}}{16}} \times {\frac{W_{x}}{16}} + 1$). 
% summarized as:
% \vspace{-3mm}
% \begin{equation}
% \label{eq-embed}
% \begin{gathered}
%      {\mathbf{I}_{\text{c}}^{z'}}, {\mathbf{I}_{\text{c}}^{x'}} = {\rm{Conv}_{4}}(\mathbf{I}_{\text{c}}^z, \mathbf{I}_{\text{c}}^x) \\
%      {\mathbf{T}_{\text{c}}^{z}}, {\mathbf{T}_{\text{c}}^{x}} = {\rm{Conv}_{2}}({\rm{MLP}}({\rm{Conv}_{2}}({\rm{MLP}}(\mathbf{I}_{\text{c}}^{z'}, \mathbf{I}_{\text{c}}^{x'}))))
% \end{gathered}
% \end{equation}
% where, ${\rm{Conv}_{4}}(.)$ and ${\rm{Conv}_{2}}(.)$ denote convolutional layers with strides of 4 and 2, respectively, used for downsampling. ${\rm{MLP}}$(.) represents a three-layer perceptron.

The input sequence $\mathbf{T}$ is fed into the backbones of both the student and teacher networks for feature extraction. Each backbone consists of a series of transformer blocks. In the student model, several feed-forward networks (FFNs) within these blocks are replaced by Token-Pooling-based MoE modules, which strengthen the student’s modeling capacity through expert collaboration and specialization.
After passing through the backbones, we obtain the student features $\mathbf{F}_{\text{s}}$ and teacher features $\mathbf{F}_{\text{t}}$. These features are further processed by their respective prediction heads to generate the final tracking results. In addition, $\mathbf{F}_{\text{s}}$ and $\mathbf{F}_{\text{t}}$ are input to the Adaptive Net, which decides whether to use the teacher features $\mathbf{F}_{\text{t}}$ and the target distribution to supervise the student. This adaptive strategy prevents redundant or misleading distillation signals, improving both training efficiency and stability.

\subsection{Token-Pooling-based MoE}
Due to the strong heterogeneity among multi-modal data, models with limited parameters often struggle to learn shared and complementary representations across modalities, which limits their modeling capability. To improve the feature extraction ability of efficient models in such scenarios, we propose a sparse expert mechanism based on token aggregation, called Token-Pooling-based Mixture-of-Experts (TP-MoE).
% Unlike traditional MoE models that use discrete gating functions to route tokens to different experts, TP-MoE adopts a similarity-based soft assignment strategy. It measures the similarity between input tokens and expert tokens and performs weighted aggregation to enable adaptive collaboration among experts. 
% Each expert focuses on inputs that are most relevant to its own feature subspace, allowing the model to capture complementary information across modalities within a shared semantic space. This design helps reduce the heterogeneity of multi-modal features and improves the overall representation quality of the model.
% In addition, the soft assignment strategy allows fully parallel computation across experts and eliminates the overhead caused by hard gating, such as token sorting and cross-expert communication. As a result, TP-MoE achieves lower latency and becomes more suitable for delay-sensitive visual tracking tasks.
Unlike traditional MoE models that use discrete gating functions for token routing, TP-MoE adopts a similarity-driven soft assignment strategy. It measures the similarity between input tokens and expert tokens and performs weighted aggregation to enable adaptive collaboration among experts. 
\begin{figure}[t]
\begin{center}
\includegraphics[width=1\linewidth]{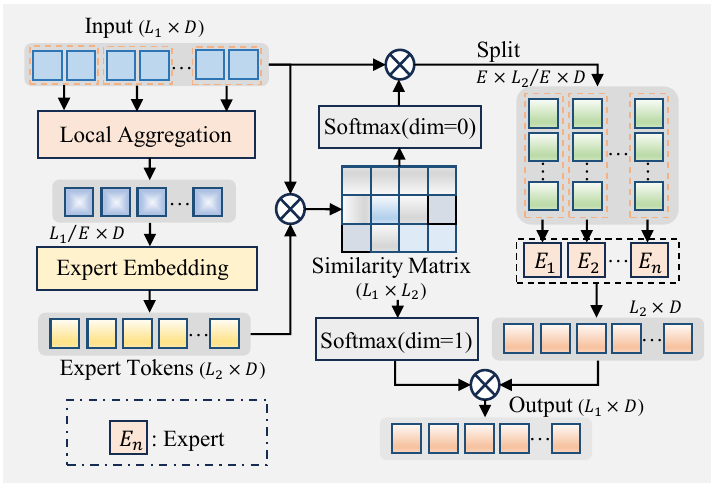}
\end{center}
\vspace{-3mm}
   \caption{TP-MoE architecture diagram.} 
\label{fig:tpmoe}
\vspace{-4mm}
\end{figure}

As shown in Figure~\ref{fig:tpmoe}, TP-MoE first enhances short-range dependency modeling through a local aggregation module. Specifically, the input tokens $\mathbf{T}_{\text{in}} \in \mathbb{R}^{L_1 \times D}$($L_1$ denotes the length of the input) are divided into ${L}_{1}/E$ subspaces, where $E$ denotes the number of experts, and average pooling is applied within each subspace. This operation strengthens local contextual relationships while preserving structural consistency among nearby tokens.
Next, the aggregated tokens are transformed into compact expert tokens $\mathbf{T}_{\text{e}} \in \mathbb{R}^{L_2 \times D}$($L_2$ denotes the length of expert tokens) through an expert embedding module, which consists of a linear projection followed by a reshape operation. 
A similarity matrix $\mathbf{S} \in \mathbb{R}^{L_1 \times L_2}$ is then computed between the input and expert tokens, and a softmax along the first dimension produces the routing weights $\mathbf{S}_{\text{a}}$. These weights acts as a continuous routing map, enabling efficient and fully parallel token–expert interactions.
Instead of relying on explicit gating or discrete routing, TP-MoE performs similarity-based soft weighting through matrix multiplication. The routing weights $\mathbf{S}_{\text{a}}$ determine how much each input token contributes to each expert, where higher similarity results in larger weights and stronger influence.
Based on $\mathbf{S}_{\text{a}}$, the input tokens are softly aggregated and sequentially grouped to form the expert inputs  $\mathbf{T}_{\text{a}} \in \mathbb{R}^{E \times \frac{L_2}{E} \times D}$. $\mathbf{T}_{\text{a}}$ contains $E$ expert groups, each with ${L}_{2}/E$ subspace tokens, ensuring that every expert focuses on distinct semantic regions.
Each expert independently processes its input to generate the expert outputs $\mathbf{O}_{\text{e}} \in \mathbb{R}^{L_2 \times D}$. Finally, these outputs are aggregated back to the input token space through another softmax weighting over the similarity matrix $\mathbf{S}$, yielding a refined and more discriminative representation $\mathbf{O} \in \mathbb{R}^{L_1 \times D}$. 
The entire process is summarized as follows:

\vspace{-2mm}
\begin{equation}
\label{eq-tpmoe}
\begin{gathered}
     {\mathbf{T}_{\text{e}}} = {\rm{Embed}}({\rm{Aggre}}(\mathbf{T}_{\text{in}})) \\
     {\mathbf{T}_{\text{a}}} = {\rm{Split}}({\rm{Softmax}}({\mathbf{T}_{\text{in}}}{\mathbf{T}_{\text{e}}^\top})^\top{\mathbf{T}_{\text{in}}})\\
     \mathbf{O}_{\text{e}} = \mathrm{Merge}\left( \left\{ \mathrm{Expert}_i(\mathbf{T}_{\text{a}}^i) \right\}_{i=1}^E \right)\\
    {\mathbf{O}} = {\rm{Softmax}}({\mathbf{T}_{\text{in}}}{\mathbf{T}_{\text{e}}^\top})\mathbf{O}_{\text{e}}
\end{gathered}
\end{equation}
where $\mathbf{T}_{\text{a}}^i$ is the input to the $i$-th expert. $\mathrm{Aggre}(\cdot)$ refers to the local aggregation, $\mathrm{Embed}(\cdot)$ is the expert embedding module, $\mathrm{Softmax}(\cdot)$ denotes the softmax activation, $\mathrm{Split}(\cdot)$ denotes sequentially partitioning tokens according to the number of experts, $\mathrm{Expert}_i(\cdot)$ represents the $i$-th expert, and $\mathrm{Merge}(\cdot)$ merges the outputs from all experts.

This attention-like soft assignment strategy can be interpreted as a subspace projection that maps input tokens onto multiple expert manifolds, where each expert focuses on the inputs most relevant to its own subspace to capture complementary semantics within a shared feature space. This mechanism encourages subspace specialization and feature diversity, thereby mitigating modality heterogeneity and enhancing representation quality. Meanwhile, the continuous routing design supports fully parallel computation and removes the overhead of hard gating, such as token sorting and inter-expert communication.
The differentiable matrix operation also stabilizes gradient propagation, resulting in lower latency and improved training stability, which is desirable for real-time visual tracking.
Through explicit local aggregation, lightweight expert embedding, and parallel soft-assignment, TP-MoE enables efficient collaboration and specialization among experts without requiring additional gating parameters or cross-expert communication. This mechanism effectively improves the model’s representation capability.
It can be flexibly integrated into the backbone by replacing the feed-forward module in transformer blocks, thus improving the model's ability to extract and fuse multi-modal features.

\subsection{Target-aware Adaptive Distillation}%Target-Aware Distillation
To further enhance model performance, we propose a distillation strategy called Target-aware Adaptive Distillation (TAD). Specifically, after the center head, the model outputs a probability distribution map of the target's center location. The teacher’s distribution map serves as a supervisory signal to guide the student via soft imitation, minimizing the divergence between the teacher and student. This encourages more accurate predictions. Additionally, feature maps from the teacher’s backbone provide auxiliary supervision to further improve the student’s ability to replicate the teacher’s representations.

\begin{figure}[t]
\begin{center}
\includegraphics[width=1\linewidth]{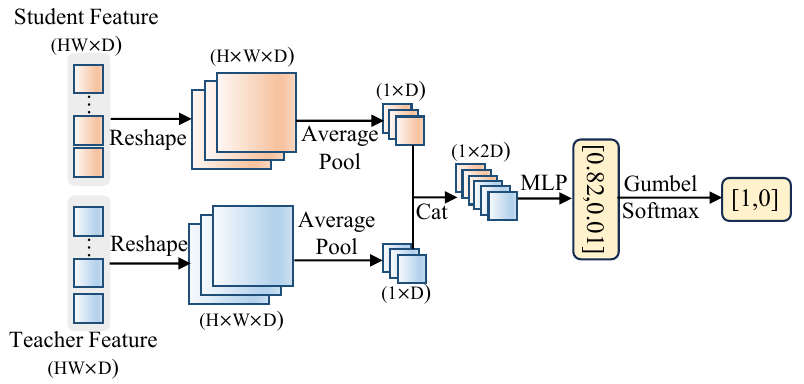}
\end{center}
\vspace{-3mm}
   \caption{Architecture of Adaptive Net.} 
\label{fig:adpnet}
\vspace{-4mm}
\end{figure}

However, for challenging samples such as those affected by occlusion, distractions, or deformation, the teacher model’s predictions may not be reliable. Directly applying distillation in these cases can transfer incorrect information to the student, introducing noisy supervision and reducing learning effectiveness. Therefore, it is necessary to prevent unreliable teacher guidance on such difficult samples to preserve the student’s learning quality on more trustworthy ones.
To address this, TAD incorporates an adaptive distillation mechanism that automatically determines whether a given sample is suitable for distillation based on its features.
The core of this mechanism is the Adaptive Net, illustrated in Figure~\ref{fig:adpnet}. It takes as input the search region feature sequence $\mathbf{T}_{\text{s}}$ from the student and $\mathbf{T}_{\text{t}}$ from the teacher. Both $\mathbf{T}_{\text{t}}$ and $\mathbf{T}_{\text{s}}$ are reshaped into 3D tensors and passed through global average pooling. The pooled features are concatenated into a fused vector $\mathbf{T}_{\text{c}}$, which is then fed into an MLP for dimensionality reduction, producing a 2D vector. This vector is converted into a one-hot vector via the Gumbel-Softmax~\cite{gumbelsoftmax} operation and output as $\mathbf{O}$ by the Adaptive Net. The value of $\mathbf{O}$ determines whether the current sample should undergo distillation, enabling fine-grained, sample-level control. 

\subsection{Training Objective}
To ensure stable training, the student and Adaptive Net are updated separately. The student’s training objective combines focal classification loss~\cite{cornernet}, GIoU~\cite{GIoU} and L1 regression losses, cross-entropy task loss~\cite{sutrack}, and distillation losses based on KL divergence and MSE, as detailed below:

\vspace{-2mm}
\begin{equation}
\begin{gathered}
\mathcal{L}_{S} = \mathcal{L}_{\text{c}}(\hat{p}_s,p) + \lambda_g \mathcal{L}_{\text{g}}(\hat{p}_s,p) + \lambda_{l_1} \mathcal{L}_{l_1}(\hat{p}_s,p)\\
+ \mathcal{L}_{t}(\hat{p}_s,p) + \alpha(\lambda_{kd}\mathcal{L}_{kd}(\hat{p}_s,\hat{p}_t)+\lambda_{f}\mathcal{L}_{f}(\hat{p}_s,\hat{p}_t))
\end{gathered}
\end{equation}
where $\mathcal{L}_{\text{c}}$, $\mathcal{L}_{\text{g}}$, $\mathcal{L}_{l_1}$, $\mathcal{L}_{\text{t}}$, $\mathcal{L}_{\text{kd}}$, and $\mathcal{L}_{\text{f}}$ denote classification, GIoU, L1, task, KL, and MSE losses, respectively. $\hat{p}_s$, $\hat{p}_t$, and $p$ are the student prediction, teacher prediction, and ground truth. Hyperparameters are $\lambda_g=2$, $\lambda_{l_1}=5$, $\lambda_{kd}=5$, and $\lambda_f=0.002$. $\alpha$ is the Adaptive Net output: $\alpha=1$ means the sample is distilled; otherwise, $\alpha=0$.

For the Adaptive Net, we adopt a surrogate prediction strategy. For each sample, it outputs a binary decision indicating whether to perform distillation. Based on this, a surrogate prediction is selected: if distillation is chosen, the teacher’s prediction serves as the target; otherwise, the student’s prediction is used. The surrogate prediction is compared with the ground truth to compute the loss, which mirrors the student’s objective but excludes distillation loss. Details are as follows:

\vspace{-2mm}
\begin{equation}
\hat{p}_a^i = 
\begin{cases} 
\hat{p}_t^i & \text{if } \alpha=1, \\
\hat{p}_s^i & \text{if } \alpha=0
\end{cases}
\vspace{-3mm}
\end{equation}

\begin{equation}
\begin{gathered}
\mathcal{L}_A = \mathcal{L}_{\text{c}}(\hat{p}_a,p) + \lambda_g \mathcal{L}_{\text{g}}(\hat{p}_a,p) + \lambda_{l_1} \mathcal{L}_{l_1}(\hat{p}_a,p) + \mathcal{L}_{\text{t}}(\hat{p}_a,p)
\end{gathered}
\end{equation}
where $\hat{p}_t^i$, $\hat{p}_s^i$, and $\hat{p}_a^i$ denote the teacher, student, and surrogate predictions for the $i$-th sample, respectively.

\section{Experiments}
\subsection{Implementation Details}

\begin{table}[t]
\centering
\caption{Details of UETrack model variants.}
\label{tab-model}
\vspace{-2mm}
\setlength{\tabcolsep}{0.5mm}{
\small
\scalebox{0.77}{
\begin{tabular}{l| c c c c c c}
\toprule
\multirow{2}{*}{Model}           & ~\multirow{2}{*}{Architecture} ~ & ~GPU~ &~CPU~ &{~AGX~}& {~Params~} & ~FLOPs~ \\
~ & & ~Speed (\emph{fps})~ & ~Speed (\emph{fps})~ & ~Speed (\emph{fps})~  & {(M)} & (G) \\
\midrule 
%\cmidrule(lr){1-1}\cmidrule(lr){2-6}
UETrack-B~~   & $[6,[6],8]$       &   $163$     &   $56$    &   $60$    &$13$    &   $3.2$\\

UETrack-S~~   & $[4,[4],4]$       &   $183$     &   $68$    &   $67$    &$9$    &   $2.5$\\

UETrack-T~~   & $[2,[2],2]$       &   $221$     &   $83$    &   $77$    &$6$   &   $1.8$\\

\bottomrule
\end{tabular}}
}
\vspace{-5mm}
\end{table}
\noindent\textbf{Model.}
UETrack is built on Fast-iTPN-T~\cite{fastitpn}, using its first $N$ layers as the backbone. The prediction head adopts a center head~\cite{ostrack}.
We develop three UETrack variants, as summarized in Table~\ref{tab-model}. In the Architecture column, $[i, [j], k]$ indicates that the backbone has $i$ layers, TP-MoE is inserted at the $j$-th layer, and $k$ experts are used. For instance, UETrack-B uses the first 6 layers of Fast-iTPN-T as the backbone, with TP-MoE at the 6th layer and 8 experts.
Table~\ref{tab-model} also reports the inference speed on 2080Ti GPU, Intel i9-14900KF CPU, and Jetson AGX Xavier, as well as model parameters and FLOPs. All models are implemented in Python 3.8.13 and PyTorch 1.13.1.

\noindent\textbf{Training.}
% We adopt SUTrack-B~\cite{sutrack} as the teacher model and keep it frozen throughout the entire training process. Only the parameters of the student model and Adaptive Net are updated.
We construct the training dataset by combining data from five common modalities, including COCO~\cite{COCO}, LaSOT~\cite{LaSOT}, GOT-10k~\cite{GOT10K}, TrackingNet~\cite{trackingnet}, VASTTrack~\cite{vasttrack}, DepthTrack~\cite{depthtrack}, VisEvent~\cite{visevent}, LasHeR~\cite{lasher}, OTB99~\cite{TNLS}, and TNL2K~\cite{TNL2K}. During training, we use RGB-X image pairs as inputs for both the template and search region, with resolutions of $224\times224$ and $112\times112$, respectively. The template and search images are generated by enlarging the bounding boxes by a factor of 2 and 4. Data augmentation includes horizontal flipping and brightness jittering.
The backbone parameters are initialized with a pretrained Fast-iTPN-T model, while the remaining ones are randomly initialized. We use the AdamW~\cite{AdamW} optimizer with an initial learning rate of $1\times10^{-5}$ for the backbone and $1\times10^{-4}$ for the rest. The weight decay is set to $1\times10^{-4}$. The model is trained for 500 epochs, with 100{,}000 samples per epoch. The learning rate is reduced by a factor of 10 after epoch 400. 
Training is conducted on two 80GB Tesla A800 GPUs with a total batch size of 128.

\noindent\textbf{Inference.}
During inference, only the student is used. To incorporate positional priors, a Hanning window penalty is applied, following standard tracking practices~\cite{ostrack}.

\begin{table*}
  \centering
    \vspace{-2mm}
      % \centering
  \caption{State-of-the-art (SOTA) comparisons on four large-scale RGB benchmarks. The top three real-time results are highlight with \textbf{\textcolor{cRed}{red}}, \textcolor{blue}{blue} and \textcolor{cGreen}{green} fonts, respectively. The top three speed across different platforms are highlighted in \textbf{bold}.}
  %The top three results are highlight with \textbf{\textcolor{cRed}{red}}, \textcolor{blue}{blue} and \textcolor{cGreen}{green} fonts, respectively.}
  \label{tab-sota-rgb}
  \vspace{-2mm}
\resizebox{0.98\linewidth}{!}{
  \setlength{\tabcolsep}{1mm}{  
  \small
  \begin{tabular}{c|l|c ccc c ccc c ccc c ccc c ccc}
    \toprule
    &\multirow{2}*{Method} &\multicolumn{3}{c}{LaSOT} & &\multicolumn{3}{c}{LaSOT$_{ext}$} & &\multicolumn{3}{c}{TrackingNet} & &\multicolumn{3}{c}{GOT-10k} & &\multicolumn{3}{c}{Speed (\textit{fps})} \\
    \cline{3-5}
    \cline{7-9}
    \cline{11-13}
    \cline{15-17}
    \cline{19-21}
   && AUC&P$_{Norm}$&P & & AUC&P$_{Norm}$&P  & & AUC&P$_{Norm}$&P & & AO&SR$_{0.5}$&SR$_{0.75}$ &&GPU &CPU& AGX\\
    \midrule[0.5pt]
    \multirow{13}*{\rotatebox{90}{Real-time}} &\cellcolor{mygray1}{UETrack-B (Ours)} &\cellcolor{mygray1}\textbf{\textcolor{cRed}{69.2}}&\cellcolor{mygray1}\textbf{\textcolor{cRed}{78.4}}&\cellcolor{mygray1}\textbf{\textcolor{cRed}{73.8}} &\cellcolor{mygray1} &\cellcolor{mygray1}\textbf{\textcolor{cRed}{48.4}}&\cellcolor{mygray1}\textbf{\textcolor{cRed}{59.0}}&\cellcolor{mygray1}\textbf{\textcolor{cRed}{54.5}} &\cellcolor{mygray1} &\cellcolor{mygray1}\textbf{\textcolor{cRed}{82.7}}&\cellcolor{mygray1}\textbf{\textcolor{cRed}{87.4}}&\cellcolor{mygray1}\textbf{\textcolor{cRed}{80.7}} &\cellcolor{mygray1} &\cellcolor{mygray1}\textbf{\textcolor{cRed}{72.6}}&\cellcolor{mygray1}\textbf{\textcolor{cRed}{82.5}}&\cellcolor{mygray1}\textbf{\textcolor{cRed}{69.8}} &\cellcolor{mygray1} &\cellcolor{mygray1}163&\cellcolor{mygray1}56&\cellcolor{mygray1}60\\
    &\cellcolor{mygray1}UETrack-S (Ours) &\cellcolor{mygray1}\textcolor{blue}{66.9}&\cellcolor{mygray1}\textcolor{blue}{76.1}&\cellcolor{mygray1}\textcolor{blue}{70.7}  &\cellcolor{mygray1} &\cellcolor{mygray1}\textcolor{blue}{47.9}&\cellcolor{mygray1}\textcolor{blue}{58.4}&\cellcolor{mygray1}\textcolor{blue}{53.9} &\cellcolor{mygray1} &\cellcolor{mygray1}\textcolor{blue}{81.4}&\cellcolor{mygray1}\textcolor{blue}{86.3}&\cellcolor{mygray1}\textcolor{blue}{78.8} &\cellcolor{mygray1} &\cellcolor{mygray1}\textcolor{blue}{71.1}&\cellcolor{mygray1}\textcolor{blue}{81.2}&\cellcolor{mygray1}\textcolor{blue}{67.0} &\cellcolor{mygray1} &\cellcolor{mygray1}183&\cellcolor{mygray1}\textbf{68}&\cellcolor{mygray1}67\\
    &\cellcolor{mygray1}UETrack-T (Ours) &\cellcolor{mygray1}{63.4}&\cellcolor{mygray1}{72.5}&\cellcolor{mygray1}{65.1}  &\cellcolor{mygray1} &\cellcolor{mygray1}{42.2}&\cellcolor{mygray1}\textcolor{cGreen}{51.5}&\cellcolor{mygray1}{46.1} &\cellcolor{mygray1} &\cellcolor{mygray1}{78.9}&\cellcolor{mygray1}{83.8}&\cellcolor{mygray1}{74.8} &\cellcolor{mygray1} &\cellcolor{mygray1}{65.3}&\cellcolor{mygray1}{75.1}&\cellcolor{mygray1}{58.4} &\cellcolor{mygray1} &\cellcolor{mygray1}\textbf{221}&\cellcolor{mygray1}\textbf{83}&\cellcolor{mygray1}\textbf{77}\\

    &AsymTrack-B~\cite{asymtrack}&\textcolor{cGreen}{64.7}&73.0&67.8  & &\textcolor{cGreen}{44.6}&-&{-} & &\textcolor{cGreen}{80.0}&\textcolor{cGreen}{84.5}&\textcolor{cGreen}{77.4} & &\textcolor{cGreen}{67.7}&\textcolor{cGreen}{76.6}&\textcolor{cGreen}{61.4} & &197&38&64\\
    
    &DyHiT~\cite{dyhit}&62.4&70.1&64.0  & &42.1&-&- & &77.9&82.2&73.8 & &62.9&71.8&55.2 & &\textbf{299}&\textbf{63}&\textbf{111}\\    
    &HiT-Base~\cite{HiT} &64.6&\textcolor{cGreen}{73.3}&\textcolor{cGreen}{68.1} & &44.1&-&- & &\textcolor{cGreen}{80.0}&84.4&77.3 & &64.0&72.1&58.1 &&175&33&61\\ 
    &MixFormerV2-S~\cite{mixformerv2}&60.6&69.9&60.4  & &43.6&-&\textcolor{cGreen}{46.2} & &75.8&81.1&70.4 & &61.9&71.7&51.3 & &\textbf{299}&47&\textbf{70}\\
    &TCTrack~\cite{tctrack}&60.5&69.3&62.4  & &-&-&- & &74.8&79.6&73.3 & &66.2&75.6&61.0 & &140&45&41\\ 
    &FEAR~\cite{fear}& 53.5&-&54.5 & &-&-&- & & -&-&- & &  61.9&72.2&- &&105&60&38\\
    &HCAT~\cite{HCAT} & 59.3&68.7&61.0 & &40.6&-&- & & 76.6&82.6&72.9  & &   65.1&76.5&56.7 &&195&45&55\\
    &E.T.Track~\cite{ETTrack}& 59.1&-&- & &-&-&- & & 75.0&80.3&70.6 &&  -&-&- &&40&47&20\\
    &LightTrack~\cite{lighttrack}& 53.8&-&53.7 & &-&-&- & & 72.5&77.8&69.5 & &  61.1&71.0&- &&128&41&36\\
    &ATOM~\cite{ATOM}& 51.5&57.6&50.5 & &37.6&45.9&43.0 & & 70.3&77.1&64.8 & &  55.6&63.4&40.2 &&83&18&22\\
    % &ECO~\cite{ECO}& 32.4&33.8&30.1 & &22.0&-&- & & 55.4&61.8&49.2 &&  31.6&30.9&11.1 &&240&15&39 \\
    \midrule[0.5pt]
     \multirow{10}*{\rotatebox{90}{Non-real-time}}
    &MCITrack-B224~\cite{mcitrack}& {{75.3}}&{{85.6}}&{{83.3}} && {{54.6}} &{{65.7}} &{{62.1}} &&  {{86.3}}&{{90.9}}&86.1 &&77.9&88.2&76.8 &&34&2&6\\
    &SUTrack-B224~\cite{sutrack}& 73.2&83.4&80.5 && 53.1 &64.2 &60.5 &&  85.7&90.3&85.1 &&77.9&87.5&78.5 &&55&6&13\\
    &MixFormerV2-B~\cite{mixformerv2}& 70.6&80.8&76.2 && 50.6 &- &56.9 &&  83.4&88.1&81.6&&73.9&-&- &&116&11&16\\
    &SeqTrack-B256~\cite{SeqTrack}& 69.9&79.7&76.3 && 49.5&60.8&56.3 &&  83.3&88.3&82.2 && 74.7&84.7&71.8 &&31&2&6\\
    &ARTrack-256~\cite{artrack}& 70.4&79.5&76.6 && 46.4&56.5&52.3 &&  84.2&88.7&83.5 && 73.5&82.2&70.9 &&39&3&6\\
    &OSTrack-256~\cite{ostrack}& 69.1&78.7&75.2 && 47.4&57.3&53.3 &&  83.1&87.8&82.0 && 71.0&80.4&68.2 &&105&11&19\\
    &Sim-B/16~\cite{simtrack}&69.3&78.5&-&&-&-&-&&82.3&86.5&-&&68.6&78.9&62.4 &&87&10&16\\
    &STARK-ST50~\cite{Stark}& 66.6&-&-  && -&-&- && 81.3&86.1&- && 68.0&77.7&62.3 &&50&7&13\\
    &TransT~\cite{transt}& 64.9&73.8&69.0 && 44.4&-&- && 81.4&86.7&83.0 && 67.1&76.8&60.9  &&63&5&13\\
    &DiMP~\cite{DiMP}& 56.9&65.0&56.7 && 39.2&47.6&45.1 && 74.0&80.1&68.7 && 61.1 &71.7&49.2 &&77&10&17 \\
  \bottomrule
\end{tabular}
}}
  \vspace{-3.5mm}
\end{table*}

\subsection{State-of-the-Art Comparisons}
We conduct a comprehensive comparison between UETrack and state-of-the-art methods across five modalities, twelve datasets, and three different hardware platforms. A tracker is defined as real-time if it runs over 20 FPS on the Jetson AGX Xavier; otherwise, it is considered non-real-time.

\begin{figure}[t]
\begin{center}
\includegraphics[width=1\linewidth]{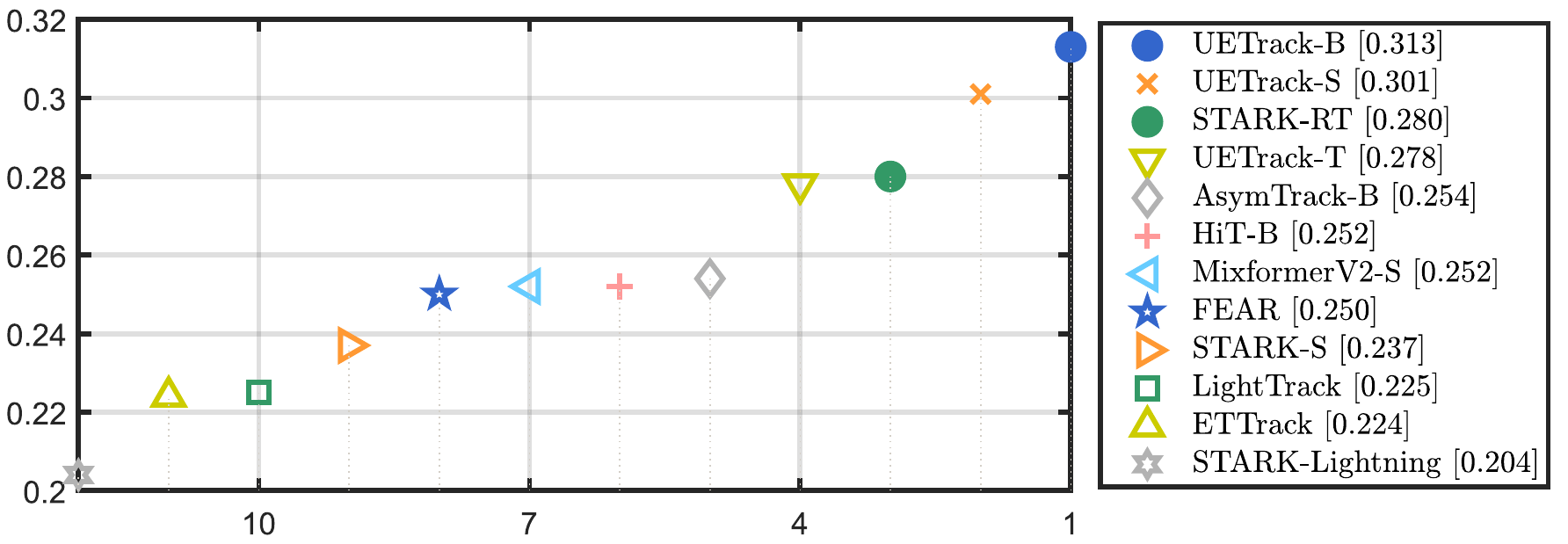}
\end{center}
\vspace{-5mm}
   \caption{EAO rank plots on VOT2021 Real-time.} 
\label{fig:vot2021}
\vspace{-6mm}
\end{figure}

\noindent\textbf{RGB-based Tracking.}
We evaluate UETrack on five RGB benchmarks, including LaSOT~\cite{LaSOT}, LaSOT$_{ext}$~\cite{lasot_journal}, TrackingNet~\cite{trackingnet}, GOT-10k~\cite{GOT10K}, and VOT2021 Real-time~\cite{vot2021}. The evaluation results are summarized in Table~\ref{tab-sota-rgb} and Figure~\ref{fig:vot2021}.
As shown, UETrack-B and UETrack-S achieve top-2 performance across all five benchmarks compared to previous real-time trackers. Specifically, UETrack-B obtains AUC scores of 69.2\%, 48.4\%, and 82.7\% on LaSOT, LaSOT$_{ext}$, and TrackingNet, respectively; an AO score of 72.6\% on GOT-10k; and an EAO score of 0.313 on the VOT2021 Real-time. These results outperform the previous best real-time tracker, AsymTrack~\cite{asymtrack}, by margins of 4.5\%, 3.8\%, 2.7\%, 4.9\%, and 0.059, respectively, setting a new state-of-the-art for real-time tracking.
Notably, compared to OSTrack~\cite{ostrack}, UETrack-B achieves higher scores on LaSOT (+0.1\%), LaSOT$_{ext}$ (+1.0\%), and GOT-10k (+1.6\%), while running significantly faster—$1.6\times$ on GPU, $5.1\times$ on CPU, and $3.2\times$ on AGX.

\begin{table}[t]\normalsize
    \caption{SOTA comparisons on depth modality.}
\label{tab-sota-rgbd-vot}
\vspace{-2mm}
  \centering
\resizebox{1\linewidth}{!}{
  \setlength{\tabcolsep}{0.5mm}{
    \small
    \begin{tabular}{c|l|cc c cc c ccc }
    \toprule
    &\multirow{2}*{Method} & \multicolumn{2}{c}{VOT-RGBD22} & & \multicolumn{2}{c}{DepthTrack} & & \multicolumn{3}{c}{Speed (\textit{fps})}\\
        \cline{3-4} \cline{6-7}  \cline{9-11}
 && EAO & Acc. & & F-score  &Re & &GPU &CPU &AGX \\
    \midrule[0.5pt]
    \multirow{7}*{\rotatebox{90}{Real-time}}
    &\cellcolor{mygray1}UETrack-B (Ours) &\cellcolor{mygray1}\textcolor{cGreen}{68.3}&\cellcolor{mygray1}\textcolor{blue}{80.8} &\cellcolor{mygray1}&\cellcolor{mygray1}\textcolor{blue}{60.6} 
    &\cellcolor{mygray1}\textcolor{blue}{61.0} 
    &\cellcolor{mygray1} &\cellcolor{mygray1}\textbf{163}&\cellcolor{mygray1}\textbf{56}&\cellcolor{mygray1}\textbf{60}\\
    &\cellcolor{mygray1}UETrack-S (Ours) &\cellcolor{mygray1}66.5 &\cellcolor{mygray1}79.9 
    &\cellcolor{mygray1}&\cellcolor{mygray1}\textcolor{cGreen}{58.9} & \cellcolor{mygray1}{58.0}&\cellcolor{mygray1} &\cellcolor{mygray1}\textbf{183}&\cellcolor{mygray1}\textbf{68}&\cellcolor{mygray1}\textbf{67}\\
    &\cellcolor{mygray1}UETrack-T (Ours) &\cellcolor{mygray1}62.5&\cellcolor{mygray1}77.4 &\cellcolor{mygray1}
    &\cellcolor{mygray1}55.7 &\cellcolor{mygray1}54.8 &\cellcolor{mygray1} &\cellcolor{mygray1}\textbf{221}&\cellcolor{mygray1}\textbf{83}&\cellcolor{mygray1}\textbf{77}\\
    &SUTrack-T~\cite{sutrack} &68.1&\textbf{\textcolor{cRed}{81.0}}&
    &\textbf{\textcolor{cRed}{61.7}} &\textbf{\textcolor{cRed}{62.1}} & &100 &23&34\\
    &EMTrack~\cite{emtrack} &\textbf{\textcolor{cRed}{69.7}}&\textcolor{cGreen}{80.6}&
    &58.3 &{58.5} & &109&29&36\\
    &CMDTrack-T9~\cite{cmdtrack}&-&-& &58.1&\textcolor{cGreen}{59.3} & &145 &51 &58 \\ 
    &ViPT-Tiny~\cite{emtrack} &\textcolor{blue}{68.5}&80.4 &
    &53.9 &53.7 & &56 &22 &20\\
    \midrule[0.1pt]
    %SeqTrackv2-L384~\cite{seqtrackv2} &74.8&82.6&91.0 & &62.3&62.6&62.5\\
    %SeqTrackv2-B256~\cite{seqtrackv2} &74.4&81.5&91.0 & &63.2&63.4&62.9 \\
    \multirow{8}*{\rotatebox{90}{Non-real-time}}
    &SeqTrackv2~\cite{seqtrackv2} &74.4&81.5& &63.2&63.4 & &23 &2 &5 \\
    &OneTracker~\cite{OneTracker} &72.7&81.9& &60.9&60.4 & &- &- &- \\
    &SDSTrack~\cite{SDSTrack} &72.8&81.2& &61.9&60.9 & & 42&3 &7 \\
    &Un-Track~\cite{untrack} &72.1&82.0& &61.0 &60.8 & & 22&4 &5 \\
    &ViPT~\cite{vipt} &72.1&81.5& &59.4 &59.6 & &55 &6 &13 \\
    % &ProTrack~\cite{protrack} &65.1&80.1& &57.8&57.3 & & -&- &- \\
    &OSTrack~\cite{ostrack} &67.6&80.3& &52.9&52.2 & &105&11&19\\
    &SPT~\cite{rgbd1k} &65.1&79.8& &53.8&54.9 & &25 &- &- \\
    %SBT-RGBD~\cite{sbt} &70.8&80.9&86.4 & &-&-&-\\
    %OSTrack~\cite{ostrack} &67.6&80.3&83.3 & &52.9&52.2&53.6\\
    &DeT~\cite{depthtrack} &65.7&76.0& &53.2&50.6 & & 37& -& -\\
    %DMTrack~\cite{vot2022} &65.8&75.8&85.1 & &-&-&-\\
    %DDiMP~\cite{vot2020} &-&-&- & &48.5&56.9&50.3\\
    %STARK-RGBD~\cite{Stark} &64.7&80.3&79.8 & &-&-&-\\
    %KeepTrack~\cite{keeptrack} &60.6&75.3&79.7 & &-&-&-\\
    %DRefine~\cite{vot2021} &59.2&77.5&76.0 & &-&-&-\\
    % &DAL~\cite{dal} &-&-&- & &42.9&36.9&51.2\\
    %ATCAIS~\cite{vot2020} &55.9&76.1&73.9 & &47.6&45.5&50.0\\
    %LTMU-B~\cite{LTMU} &-&-&- & &46.0&41.7&51.2\\
    %GLGS-D~\cite{vot2020} &-&-&- & &45.3&36.9&58.4\\
    %LTDSEd~\cite{VOT2019} &-&-&- & &40.5&38.2&43.0\\
    %Siam-LTD~\cite{vot2020} &-&-&- & &37.6&34.2&41.8\\
    %SiamM-Ds~\cite{VOT2019} &-&-&- & &33.6&26.4&46.3\\
    %CA3DMS~\cite{ca3dms} &-&-&- & &22.3&22.8&21.8\\
    %DiMP~\cite{DiMP} &54.3&70.3&73.1 & &-&-&-\\
    %ATOM~\cite{ATOM} &50.5&59.8&68.8 & &-&-&-\\
    \bottomrule
    \end{tabular}
    }
  }
  \vspace{-6mm}
\end{table}

\noindent\textbf{RGB-Depth Tracking.}
UETrack delivers strong performance on RGB-Depth tasks while maintaining high inference speed. As shown in Table~\ref{tab-sota-rgbd-vot}, on the VOT-RGBD22 benchmark~\cite{vot2022}, UETrack-B achieves an EAO of 68.3\%, surpassing SUTrack-T by 0.2\%, and runs $1.6\times$, $2.4\times$, and $1.8\times$ faster on GPU, CPU, and AGX, respectively.
On DepthTrack~\cite{depthtrack}, UETrack-B achieves an F-score of 60.6\%. It outperforms EMTrack by 2.3\%, with speed gains of $1.5\times$, $1.9\times$, and $1.7\times$ on GPU, CPU, and AGX, respectively. Compared to ViPT, UETrack-B achieves a 1.2\% higher F-score, and runs $3.0\times$, $9.3\times$, and $4.6\times$ faster on GPU, CPU, and AGX, respectively.

\noindent\textbf{RGB-Thermal Tracking.}
UETrack achieves best real-time performance on LasHeR~\cite{lasher} and RGBT234~\cite{rgbt234}, as shown in Table~\ref{tab-sota-rgbt}. UETrack-B records 55.5\% AUC on LasHeR and 64.2\% MSR on RGBT234, surpassing SUTrack-T by 1.6\% and 0.4\%, respectively. Compared to the non-real-time SDSTrack, UETrack-B improves by 2.4\% on LasHeR and 1.7\% on RGBT234, while running $3.9\times$, $18.7\times$, and $8.6\times$ faster on GPU, CPU, and AGX, respectively.

\begin{table}[t]\Large
    \caption{SOTA comparisons on thermal modality.}
\label{tab-sota-rgbt}
\vspace{-2mm}
  \centering
\resizebox{1\linewidth}{!}{
  \setlength{\tabcolsep}{0.5mm}{
    \small
    \begin{tabular}{c|l|cc c cc c ccc}
    \toprule
    &\multirow{2}*{Method} & \multicolumn{2}{c}{LasHeR} & & \multicolumn{2}{c}{RGBT234} &&\multicolumn{3}{c}{Speed (\textit{fps})}  \\
        \cline{3-4} \cline{6-7} \cline{9-11}
 && AUC & P & &MSR &MPR &&GPU &CPU &AGX\\
    \midrule[0.5pt]
    \multirow{7}*{\rotatebox{90}{Real-time}}
    &\cellcolor{mygray1}UETrack-B (Ours) &\cellcolor{mygray1}\textbf{\textcolor{cRed}{55.5}}&\cellcolor{mygray1}\textbf{\textcolor{cRed}{69.1}} &\cellcolor{mygray1} &\cellcolor{mygray1}\textbf{\textcolor{cRed}{64.2}}&\cellcolor{mygray1}\textbf{\textcolor{cRed}{86.7} }&\cellcolor{mygray1}&\cellcolor{mygray1}\textbf{163}&\cellcolor{mygray1}\textbf{56}&\cellcolor{mygray1}\textbf{60} \\
    &\cellcolor{mygray1}UETrack-S (Ours) &\cellcolor{mygray1}{53.2}&\cellcolor{mygray1}\textcolor{cGreen}{66.4} &\cellcolor{mygray1} &\cellcolor{mygray1}\textcolor{cGreen}{62.2}&\cellcolor{mygray1}\textcolor{cGreen}{84.4}&\cellcolor{mygray1}&\cellcolor{mygray1}\textbf{183}&\cellcolor{mygray1}\textbf{68}&\cellcolor{mygray1}\textbf{67} \\
    &\cellcolor{mygray1}UETrack-T (Ours) &\cellcolor{mygray1}{48.2}&\cellcolor{mygray1}59.8 &\cellcolor{mygray1} &\cellcolor{mygray1}59.3&\cellcolor{mygray1}81.3&\cellcolor{mygray1}&\cellcolor{mygray1}\textbf{221}&\cellcolor{mygray1}\textbf{83}&\cellcolor{mygray1}\textbf{77} \\
    &SUTrack-T~\cite{sutrack} &\textcolor{blue}{53.9}&\textcolor{blue}{66.7} & &\textcolor{blue}{63.8}&\textcolor{blue}{85.9}&&100 &23&34 \\
    &EMTrack~\cite{emtrack} &\textcolor{cGreen}{53.3} &65.9 &&60.1 &81.8 &&109&29&36 \\
    &CMDTrack-T9~\cite{cmdtrack} &52.8 &65.4 & &59.8 &83.1 & &145&51&58\\
    &ViPT-Tiny~\cite{emtrack}  &47.5 &58.5 &&58.8 &80.0 &&56 &22 &20\\
    \midrule[0.1pt]
    %SeqTrackv2-L384~\cite{seqtrackv2} &61.0&76.7 & &68.0&91.3\\
    %SeqTrackv2-B256~\cite{seqtrackv2} &55.8&70.4 & &64.7&88.0 \\
    \multirow{9}*{\rotatebox{90}{Non-real-time}}
    &SeqTrackv2~\cite{seqtrackv2} &55.8&70.4 & &64.7&88.0&& 23&2 &5 \\
    &OneTracker~\cite{OneTracker} &53.8&67.2 & &64.2&85.7&& -& -& -\\
    &SDSTrack~\cite{SDSTrack} &53.1&66.5 & &62.5&84.8&& 42&3 &7 \\
    &Un-Track~\cite{untrack} &51.3&54.6 & &62.5&84.2&& 22&4 &5 \\
    &ViPT~\cite{vipt} &52.5&65.1 & &61.7&83.5&& 55&6 &13 \\
    &ProTrack~\cite{protrack} &42.0&53.8 & &59.9&79.5&& -&- &- \\
    &BAT~\cite{bat} &56.3&70.2 & &64.1&86.8&& 56&3 &7 \\
    &TBSI~\cite{tbsi} &55.6&69.2 & &63.7&87.1&& 42&2 &5 \\
    &TATrack~\cite{TATrack-T} &56.1&70.2 & &64.4&87.2&& 26&- &- \\
    % &OSTrack~\cite{ostrack} &41.2&51.5 & &54.9&72.9&&105 &11 &19 \\
    %TransT~\cite{transt} &39.4&52.4 &&\\
    % &APFNet~\cite{apfnet} &36.2&50.0 & &57.9&82.7&& & & \\
    % &JMMAC~\cite{jmmac} &-&- & &57.3&79.0&& & & \\
    % &CMPP~\cite{cmpp} &-&- & &57.5&82.3&& & & \\
    %STARK~\cite{Stark} &36.1&44.9 &&\\
    %mfDiMP~\cite{mfdimp} &34.3&44.7 & &42.8&64.6\\
    %DAPNet~\cite{dapnet} &31.4&43.1 & &-&-\\
    % &CAT~\cite{cat} &31.4&45.0 & &56.1&80.4&& & & \\
    % &HMFT~\cite{vtuav} &31.3&43.6 & &-&-&& & & \\
    % MaCNet~\cite{macnet} &-&- & &55.4&79.0&& & & \\
    % &FANet~\cite{fanet} &30.9&44.1 & &55.3&78.7&& & & \\
    % &DAFNet~\cite{dafnet} &-&- & &54.4&79.6 && & & \\
    %SGT~\cite{sgt} &25.1&36.5 & &47.2&72.0 \\
    %SGT++~\cite{} &25.1&36.5 & &-&-\\
    \bottomrule
    \end{tabular}
    }
  }
  \vspace{-4mm}
\end{table}

\begin{table}[t]\tiny
    \caption{SOTA comparisons on event modality.}
\label{tab-sota-rgbe}
\vspace{-2mm}
  \centering
\resizebox{1\linewidth}{!}{
  \setlength{\tabcolsep}{1.5mm}{
    \small
    \begin{tabular}{c|l|cc c ccc}
    \toprule
    &\multirow{2}*{Method} & \multicolumn{2}{c}{VisEvent} &&\multicolumn{3}{c}{Speed (\textit{fps})}\\
        \cline{3-4}  \cline{6-8}
 && AUC & P &&GPU &CPU &AGX\\
    \midrule[0.5pt]
    \multirow{7}*{\rotatebox{90}{Real-time}}
    &\cellcolor{mygray1}UETrack-B (Ours) &\cellcolor{mygray1}\textbf{\textcolor{cRed}{59.2}}&\cellcolor{mygray1}\textbf{\textcolor{cRed}{76.2}} &\cellcolor{mygray1} &\cellcolor{mygray1}\textbf{163}&\cellcolor{mygray1}\textbf{56}&\cellcolor{mygray1}\textbf{60}\\
    &\cellcolor{mygray1}UETrack-S (Ours) &\cellcolor{mygray1}58.0 &\cellcolor{mygray1}\textcolor{cGreen}{75.1}  &\cellcolor{mygray1} &\cellcolor{mygray1}\textbf{183}&\cellcolor{mygray1}\textbf{68}&\cellcolor{mygray1}\textbf{67}\\
    &\cellcolor{mygray1}UETrack-T (Ours) &\cellcolor{mygray1}54.4&\cellcolor{mygray1}71.8 &\cellcolor{mygray1} &\cellcolor{mygray1}\textbf{221}&\cellcolor{mygray1}\textbf{83}&\cellcolor{mygray1}\textbf{77}\\
    &SUTrack-T~\cite{sutrack} &\textcolor{blue}{58.8}&\textcolor{blue}{75.7} & &100 &23&34\\
    &EMTrack~\cite{emtrack} &\textcolor{cGreen}{58.4}&72.4& &109&29&36\\
    &CMDTrack-T9~\cite{cmdtrack}&57.9&72.4 & &145 &51 &58\\
    &ViPT-Tiny~\cite{emtrack} &55.8&69.8 & &56 &22 &20\\
     \midrule[0.1pt]
    %SeqTrackv2-L384~\cite{seqtrackv2} &63.4&80.0\\
    %SeqTrackv2-B256~\cite{seqtrackv2} &61.2&78.2 \\
    \multirow{8}*{\rotatebox{90}{Non-real-time}}
    &SeqTrackv2~\cite{seqtrackv2} &61.2&78.2 && 23&2 &5 \\
    &OneTracker~\cite{OneTracker} &60.8&76.7 &&- &- &- \\
    &SDSTrack~\cite{SDSTrack} &59.7 &76.7 && 42&3 &7 \\
    &Un-Track~\cite{untrack} &58.9 &75.5 && 22&4 &5 \\
    &ViPT~\cite{vipt} &59.2&75.8 && 55&6 &13 \\
    &ProTrack~\cite{protrack} &47.1&63.2 && -&- &- \\
    &OSTrack~\cite{ostrack} &53.4&69.5 &&105 &11 &19 \\
    % &STARK~\cite{Stark} &44.6&61.2 &&50 &7 &13 \\
    % &TransT~\cite{transt} &47.4&65.0 &&63 &5 &13 \\
    %LTMU\_E~\cite{LTMU} &45.9&65.5 \\
    %PrDiMP\_E~\cite{PrDiMP} &45.3&64.4 \\
    %STARK\_E~\cite{Stark} &44.6&61.2 \\
    %MDNet\_E~\cite{MDNet} &42.6&66.1 \\
    %SiamCar\_E~\cite{Stark} &42.0&59.9 \\
    %VITAL\_E~\cite{VITAL} &41.5&64.9 \\
    %ATOM\_E~\cite{ATOM} &41.2&60.8 \\
    %SiamBAN\_E~\cite{SiamBAN} &40.5&59.1 \\
    %SiamMask\_E~\cite{SiamMask} &36.9&56.2 \\
    \bottomrule
    \end{tabular}
    }
  }
  \vspace{-5mm}
\end{table}

\noindent\textbf{RGB-Event Tracking.}
As shown in Table~\ref{tab-sota-rgbe}, UETrack achieves a new real-time state-of-the-art on VisEvent~\cite{visevent}. Specifically, UETrack-B obtains an AUC score of 59.2\%, surpassing the previous real-time trackers SUTrack-T and EMTrack by 0.4\% and 0.8\%, respectively.

\noindent\textbf{RGB-Language Tracking.}
UETrack also demonstrates competitive performance on the Language modality. As shown in Table~\ref{tab-sota-rgbl}, UETrack-B achieves an AUC score of 58.0\% on TNL2K~\cite{TNL2K}, surpassing SeqTrackv2 by 0.5\%, while running $7.1\times$, $28\times$, and $12\times$ faster on GPU, CPU, and AGX, respectively. On OTB99~\cite{TNLS}, UETrack-B, UETrack-S, and UETrack-T achieve AUC scores of 61.3\%, 63.1\%, and 64.8\%, respectively. 

\begin{table}[t]\Large
    \caption{SOTA comparisons on language modality.}
\label{tab-sota-rgbl}
\vspace{-2mm}
  \centering
\resizebox{1\linewidth}{!}{
  \setlength{\tabcolsep}{0.5mm}{
    \small
    \begin{tabular}{c|l|cc c cc c ccc}
    \toprule
    &\multirow{2}*{Method} & \multicolumn{2}{c}{TNL2K} & & \multicolumn{2}{c}{OTB99} &&\multicolumn{3}{c}{Speed (\textit{fps})}  \\
        \cline{3-4} \cline{6-7} \cline{9-11}
 && AUC & P & &AUC &P &&GPU &CPU &AGX\\
    \midrule[0.5pt]
    \multirow{4}*{\rotatebox{90}{Real-time}}
    &\cellcolor{mygray1}UETrack-B (Ours) &\cellcolor{mygray1}\textcolor{blue}{58.0}&\cellcolor{mygray1}\textcolor{blue}{60.3} &\cellcolor{mygray1} &\cellcolor{mygray1}{61.3}&\cellcolor{mygray1}{79.7} &\cellcolor{mygray1}&\cellcolor{mygray1}\textbf{163}&\cellcolor{mygray1}\textbf{56}&\cellcolor{mygray1}\textbf{60} \\
    &\cellcolor{mygray1}UETrack-S (Ours) &\cellcolor{mygray1}\textcolor{cGreen}{57.0}&\cellcolor{mygray1}\textcolor{cGreen}{58.4} &\cellcolor{mygray1} &\cellcolor{mygray1}\textcolor{cGreen}{63.1}&\cellcolor{mygray1}\textcolor{cGreen}{81.9}&\cellcolor{mygray1}&\cellcolor{mygray1}\textbf{183}&\cellcolor{mygray1}\textbf{68}&\cellcolor{mygray1}\textbf{67} \\
    &\cellcolor{mygray1}UETrack-T (Ours) &\cellcolor{mygray1}{54.4}&\cellcolor{mygray1}54.4 &\cellcolor{mygray1} &\cellcolor{mygray1}\textcolor{blue}{64.8}&\cellcolor{mygray1}\textcolor{blue}{84.7}&\cellcolor{mygray1}&\cellcolor{mygray1}\textbf{221}&\cellcolor{mygray1}\textbf{83}&\cellcolor{mygray1}\textbf{77} \\
    &SUTrack-T~\cite{sutrack} &\textbf{\textcolor{cRed}{60.9}}&\textbf{\textcolor{cRed}{62.3}} & &\textbf{\textcolor{cRed}{67.4}}&\textbf{\textcolor{cRed}{88.6}}&&100 &23&34 \\
    \midrule[0.1pt]
    %SeqTrackv2-L384~\cite{seqtrackv2} &61.0&76.7 & &68.0&91.3\\
    %SeqTrackv2-B256~\cite{seqtrackv2} &55.8&70.4 & &64.7&88.0 \\
    \multirow{9}*{\rotatebox{90}{Non-real-time}}
    &SeqTrackv2~\cite{seqtrackv2} &57.5&59.7 & &71.2&93.9 &&23 &2 &5 \\
    &OneTracker~\cite{OneTracker} &58.0&59.1 & &69.7&91.5 &&- &- &- \\
    &UVLTrack-B~\cite{uvltrack} &63.1 &66.7 &&69.3&89.9 && 52&4 &6 \\
    &CiteTracker~\cite{citeTrack} &57.7 &59.6 &&69.6&85.1 &&22 &2 &4 \\
    &JointNLT~\cite{JointNLT}	&56.9 &58.1 & &65.3&85.6 &&39 &3 &5 \\
    &DecoupleTNL\cite{DecoupleTNL}&56.7&56.0 & &{73.8}&{94.8} &&32 &- &- \\
    &Zhao \emph{et al.}~\cite{zhao2023transformervision-languagetracking}	&56.0&- & &69.9&91.2 &&36 &- & -\\
    %VLT$_{TT}$~\cite{VLTTT}	&53.1&53.3 & &76.4 &93.1 \\
    %CapsuleTNL~\cite{CapsuleTNL}	&- &- & &71.1&92.4\\
    % &Li \emph{et al.}\cite{CTRTNL} &44.0&45.0 & &69.0&91.0 && & & \\
    % &TNL2K-2~\cite{TNL2K}	&42.0&42.0 & &68.0&88.0 && & & \\
    &SNLT~\cite{SNLT}	&27.6&41.9 & &66.6&80.4 && 50&- &- \\
    %GTI~\cite{GTI} &-&- & &58.1&73.2\\
    % &TransVG~\cite{transvg} &26.1&28.9 & &-&- && & & \\
    % &Feng \emph{et al.}~\cite{feng2019robust}	&25.0&27.0 & &67.0&73.0 && & & \\
    &RTTNLD~\cite{RTTNLD}	&25.0&27.0 & &61.0&79.0 &&30 &- &- \\
    %SGT~\cite{sgt} &25.1&36.5 & &47.2&72.0 \\
    %SGT++~\cite{} &25.1&36.5 & &-&-\\
    \bottomrule
    \end{tabular}
    }
  }
  \vspace{-3mm}
\end{table}

\noindent\textbf{Speed comparison.}
We compare the tracking speed on three platforms. UETrack consistently achieves better speed-accuracy trade-offs than previous trackers. For example, the fastest variant, UETrack-T, runs at 221 FPS on GPU, 83 FPS on CPU, and 77 FPS on AGX, outperforming most RGB-only trackers.
In RGB-X tasks, multi-modal processing typically introduces extra latency, slowing down existing trackers. However, UETrack significantly boosts multi-modal tracking speed. Compared to the unified SUTrack-T, UETrack-T achieves $2.2\times$, $3.6\times$, and $2.3\times$ higher speed on GPU, CPU, and AGX, respectively.
Overall, UETrack runs fast on all three platforms and supports five modalities, validating its practicality and versatility.

\begin{table}[t]
\centering
\caption{Ablation Study. $\Delta$ denotes the performance change (averaged over benchmarks) compared with the baseline. The speed is measured on the AGX.
}

\label{tab-ablation}
\vspace{-1.5mm}
\small
\resizebox{1\linewidth}{!}{
\setlength{\tabcolsep}{0.5mm}{
\begin{tabular}{l|c|cccccc|c}
\toprule
\# &Method &LaSOT &DepthTrack  &RGBT234 &VisEvent &TNL2K  & Speed&$\Delta$\\
\midrule
\rowcolor{mygray1}1 &Baseline &68.5 &59.4 &63.1 &57.9 &57.1 &60 &--\\
\midrule
2 &W/o TP-MoE &67.9 &58.1 &62.2 &57.3 &56.4 &63 &\textbf{-0.8}\\  
3 &Gate-MoE &68.7 &59.1 &62.9 &57.6 &56.7 &39 &\textbf{-0.2}\\  
4 &W/o Local Agg. &68.0 &59.3 &63.2 &57.0 &56.8 &61 &\textbf{-0.3}\\   
\midrule
5 &4 Experts &68.0 &59.5 &62.4 &57.4 &56.7 &60 &\textbf{-0.4}\\ 
6 &16 Experts &68.2 &58.6 &62.2 &57.5 &56.6 &60 &\textbf{-0.6}\\ 
7 &32 Experts &68.5 &59.3 &62.4 &57.1 &56.4 &59 &\textbf{-0.5}\\ 
\midrule
8 &Last 2 Layers &67.7 &59.3 &63.2 &57.6 &57.2 &53 &\textbf{-0.2}\\  
9 &Last 3 Layers &67.5 &58.1 &62.7 &57.3 &57.4 &48 &\textbf{-0.6}\\
10 &Even Layers &67.2 &59.2 &62.2 &57.5 &57.0 &48 &\textbf{-0.6}\\
\midrule
11 & + KL &68.7 &59.7 &63.7 &58.0 &57.2 &60 &\textbf{+0.3}\\
12 & + Feature Mim. &69.0 &60.1 &63.7 &58.2 &57.4 &60 &\textbf{+0.5}\\
13 & + Adaptive &69.2 &60.6 &64.2 &59.2 &58.0 &60 &\textbf{+1.0}\\
\bottomrule
\end{tabular}
}}
\vspace{-4mm}
\end{table}
\subsection{Ablation and Analysis}
As shown in Table~\ref{tab-ablation}, we conduct extensive ablation experiments to validate the effectiveness of the proposed TP-MoE and TAD.
In Table~\ref{tab-ablation}, models \#1 to \#10 are all trained without TAD. The baseline model (\#1) is UETrack-B, which incorporates TP-MoE but does not use TAD.

\begin{figure}[t]
\begin{center}
\includegraphics[width=1\linewidth]{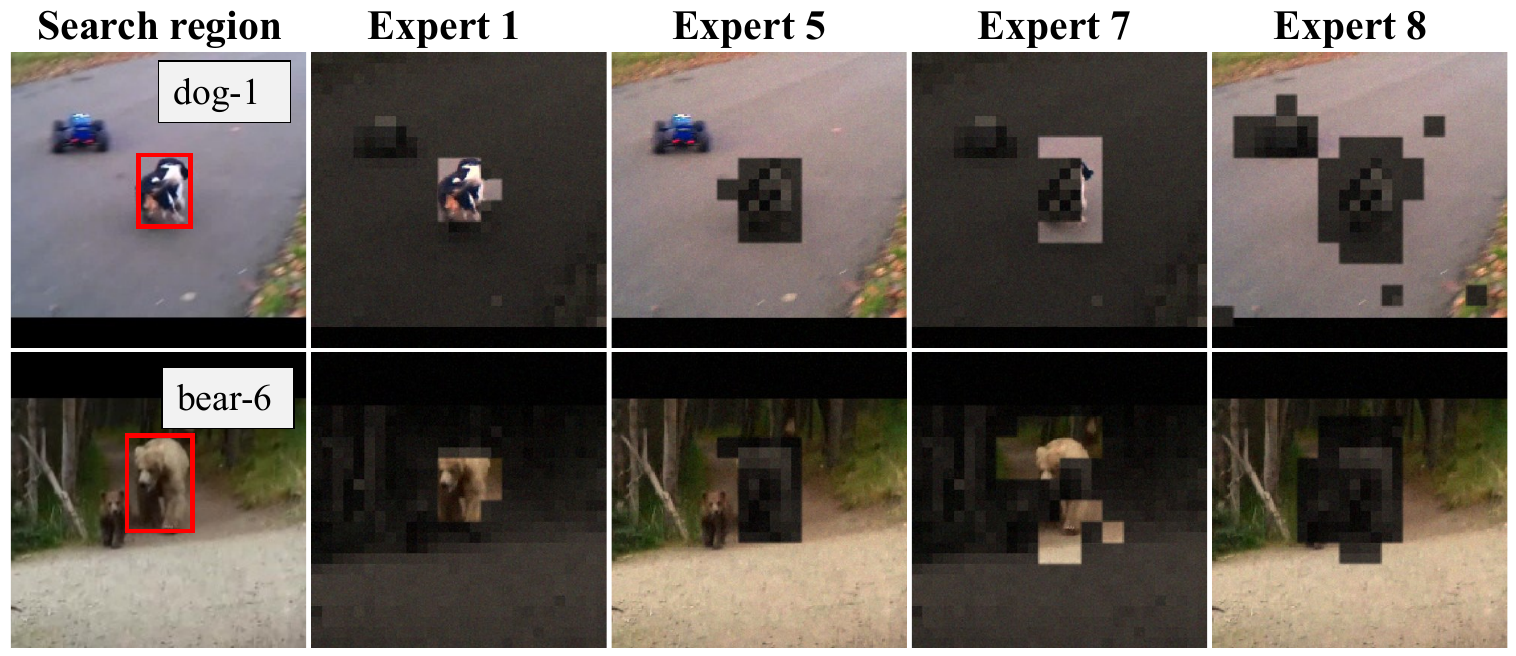}
\end{center}
\vspace{-5mm}
   \caption{Visualization of attention distributions of TP-MoE experts. The bright regions denote the attended areas. Each expert focuses on distinct spatial regions.} 
\label{fig:tpmoe-vis}
\vspace{-6mm}
\end{figure}

\noindent\textbf{Necessity of TP-MoE.}
To verify the effectiveness of TP-MoE, we conduct three groups of experiments. In \#2, TP-MoE is entirely removed. In \#3, it is replaced by a gated MoE that assigns tokens through a gating mechanism. In \#4, the local aggregation process within TP-MoE is removed.
As observed, removing TP-MoE (\#2) leads to performance drops across multiple datasets, with an average decrease of 0.8\%. Replacing TP-MoE with the gated MoE (\#3) also causes a slight drop of 0.2\% on average. Moreover, due to the time-consuming gating mechanism in the gated MoE, the model speed drops significantly—by 21 FPS compared to the baseline. When the local aggregation is removed (\#4), the model shows an average accuracy decrease of 0.3\%.
These results demonstrate the necessity of TP-MoE. It enhances the model’s ability to process multi-modal inputs, while the similarity-driven soft assignment replaces explicit gating to maintain high efficiency.

\noindent\textbf{Number of Experts.}
The number of experts used in TP-MoE is a critical parameter. Too few experts can limit the model's representation capacity, while too many may introduce redundancy.
We evaluate this factor by varying the number of experts, as shown in Table~\ref{tab-ablation}, entries \#5, \#6, and \#7, where we use 4, 16, and 32 experts, respectively. The baseline model uses 8 experts by default.
As the results show, using 4, 16, and 32 experts leads to average performance drops of 0.4\%, 0.6\%, and 0.5\%, respectively.

\noindent\textbf{Insertion Layer of TP-MoE.}
We further explore where to insert TP-MoE within the backbone. As shown in Table~\ref{tab-ablation}, entries \#8, \#9, and \#10 correspond to inserting TP-MoE in the last two layers, the last three layers, and all even-numbered layers, respectively. The baseline model inserts TP-MoE only in the last layer.
The results show that inserting TP-MoE in the last two layers, last three layers, and even-numbered layers leads to average performance drops of 0.2\%, 0.6\%, and 0.6\%, respectively.
We attribute this to the fact that semantic features in the deeper layers are more stable and abstract, making them more suitable for expert specialization. In contrast, inserting TP-MoE into earlier layers may disrupt the still-forming feature representations, causing interference and performance degradation.

\noindent\textbf{Effectiveness of TAD.}
To validate the effectiveness of the proposed TAD, we perform ablation studies on its individual components. As shown in Table~\ref{tab-ablation}, entry \#11 introduces KL divergence supervision based on the target distribution. Entry \#12 further adds feature-level supervision, and entry \#13 incorporates adaptive distillation.
The results show that introducing KL divergence improves average performance by 0.3\%. Adding feature distillation further increases the gain to 0.5\%. Finally, incorporating adaptive distillation leads to a total improvement of 1.0\% over the baseline.
These results demonstrate the effectiveness of TAD in efficiently transferring knowledge from teacher to student.

\begin{figure}[t]
\begin{center}
\includegraphics[width=1\linewidth]{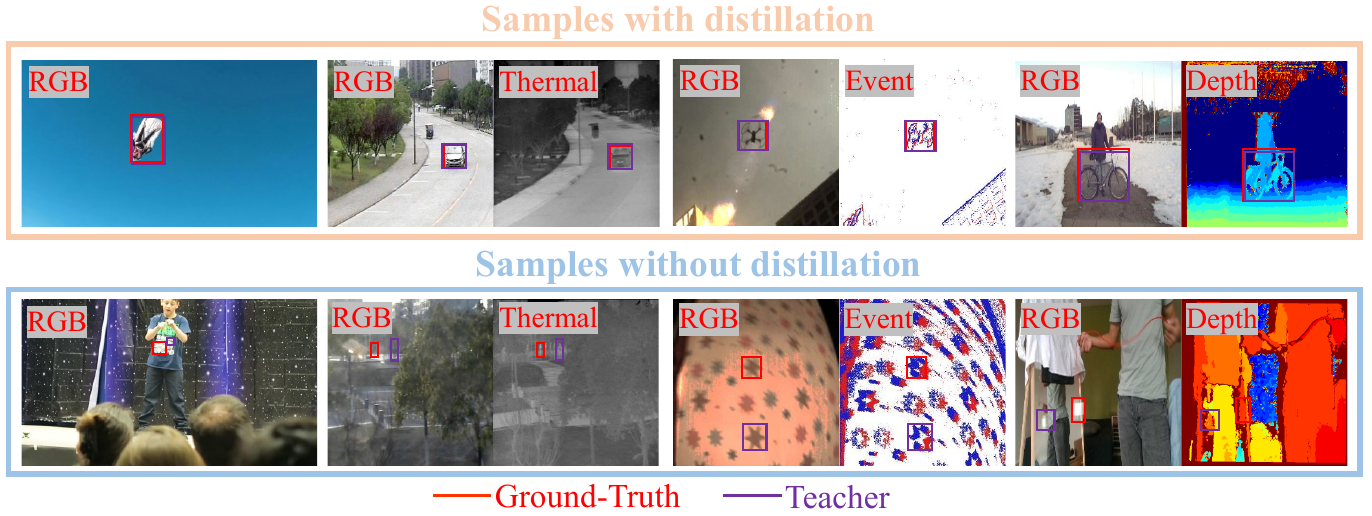}
\end{center}
\vspace{-5mm}
   \caption{Visualization of adaptive distillation decisions made by TAD across different modalities.} 
\label{fig:tad-vis}
\vspace{-6mm}
\end{figure}

\noindent\textbf{Visualization.}
We visualize the attention distributions of several experts in TP-MoE, as shown in Figure~\ref{fig:tpmoe-vis}. Each expert focuses on different regions. Specifically, Expert 1 attends to the object center, Expert 5 and Expert 8 focus on the background, while Expert 7 concentrates on the object contour. Such collaboration and clear division of attention enable experts to learn complementary representations, thereby enhancing the model’s feature modeling capability.
We also visualize the distillation decisions of TAD, as shown in Figure~\ref{fig:tad-vis}. When the scene contains challenges such as blur, occlusion, or deformation, the teacher model often makes inaccurate predictions, and TAD skips distillation for these unreliable samples. This demonstrates the effectiveness of TAD, as it prevents the student from being misled by incorrect supervision.

\section{Conclusion}
We propose UETrack, a unified and efficient tracking framework trained once and deployed across five modality-specific tasks. To improve multi-modal modeling and versatility, UETrack introduces a Token-Pooling-based MoE module for expert collaboration and a Target-aware Adaptive Distillation strategy to selectively transfer knowledge from teacher models. These designs broaden the scope of efficient trackers while improving speed and practicality in multi-modal tracking. Extensive experiments show UETrack achieves strong versatility and reliability across scenarios. We hope UETrack bridges research and real-world use, promoting practical multi-modal tracking. \\
\textit{Acknowledgements} The paper is supported in part by National Natural Science Foundations of China (no. U23A20384 and no. 62402084), in part Fundamental Scientific Research Funding of the Central Universities of China (DUTZD25225), Liaoning Provincial Science and Technology Joint Program Project (2024011188-JH2/1026), China Postdoctoral Science Foundation (no. 2024M750319).

% versatility

% UETrack significantly broadens the application scope of efficient trackers while greatly improving the speed and practicality of multi-modal tracking.
% Extensive experiments demonstrate that UETrack achieves strong generalization and reliability across different scenarios. We hope UETrack can serve as a bridge between research and real-world applications, and promote the practical development of multi-modal object tracking.

% \clearpage

{
    \small
    \bibliographystyle{ieeenat_fullname}
    \bibliography{main}

@String(IJCV = {Int. J. Comput. Vis.})

@String(CVPR= {IEEE Conf. Comput. Vis. Pattern Recog.})

@String(ICCV= {Int. Conf. Comput. Vis.})

@String(ECCV= {Eur. Conf. Comput. Vis.})

@String(NIPS= {Adv. Neural Inform. Process. Syst.})

@String(ICPR = {Int. Conf. Pattern Recog.})

@String(TIP  = {IEEE Trans. Image Process.})

@String(TMM  = {IEEE Trans. Multimedia})

@String(ACMMM= {ACM Int. Conf. Multimedia})

@String(ICLR = {Int. Conf. Learn. Represent.})

@String(PR   = {Pattern Recognition})

@String(AAAI = {AAAI})

@String(CVM = {Computational Visual Media})

@String(TPAMI  = {IEEE TPAMI})

@String(IJCV  = {IJCV})

@String(CVPR  = {CVPR})

@String(ICCV  = {ICCV})

@String(ECCV  = {ECCV})

@String(NIPS  = {NeurIPS})

@String(ICPR  = {ICPR})

@String(TIP   = {IEEE TIP})

@String(TCSVT = {IEEE TCSVT})

@String(TMM   =	{IEEE TMM})

@String(ACMMM = {ACM MM})

@String(ICLR  = {ICLR})

@String(PR = {PR})

@String(TCYB = {IEEE TCYB})

@article{switchtransformer,
  title={Switch Transformers: Scaling to Trillion Parameter Models with Simple and Efficient Sparsity},
  author={Fedus, William and Zoph, Barret and Shazeer, Noam},
  journal={JMLR},
  pages={1--39},
  year={2022}
}

@inproceedings{transt,
  title={Transformer Tracking},
  author={Chen, Xin and Yan, Bin and Zhu, Jiawen and Wang, Dong and Yang, Xiaoyun and Lu, Huchuan},
  booktitle=CVPR,
  pages={8126--8135},
  year={2021}
}

@inproceedings{seqtrack,
  title={SeqTrack: Sequence to Sequence Learning for Visual Object Tracking},
  author={Chen, Xin and Peng, Houwen and Wang, Dong and Lu, Huchuan and Hu, Han},
  booktitle=CVPR,
  pages={14572--14581},
  year={2023}
}

@inproceedings{glam,
  title={GLaM: Efficient Scaling of Language Models with Mixture-of-Experts},
  author={Du, Nan and Huang, Yanping and Dai, Andrew M and Tong, Simon and Lepikhin, Dmitry and Xu, Yuanzhong and Krikun, Maxim and Zhou, Yanqi and Yu, Adams Wei and Firat, Orhan and others},
  booktitle={ICML},
  pages={5547--5569},
  year={2022},
}

@inproceedings{TNL2K,
  title={Towards More Flexible and Accurate Object Tracking with Natural Language: Algorithms and Benchmark},
  author={Wang, Xiao and Shu, Xiujun and Zhang, Zhipeng and Jiang, Bo and Wang, Yaowei and Tian, Yonghong and Wu, Feng},
  booktitle=CVPR,
  pages={13763--13773},
  year={2021}
}

@inproceedings{vot2022,
	title={The Tenth Visual Object Tracking VOT2022 Challenge Results},
	author={Kristan, Matej and Leonardis, Ale{\v{s}} and Matas, Ji{\v{r}}{\'\i} and Felsberg, Michael and Pflugfelder, Roman and K{\"a}m{\"a}r{\"a}inen, Joni-Kristian and Chang, Hyung Jin and Danelljan, Martin and Zajc, Luka {\v{C}}ehovin and Luke{\v{z}}i{\v{c}}, Alan and others},
	booktitle={ECCVW},
	pages={431--460},
	year={2023}
}

@inproceedings{vot2021,
	title={The Ninth Visual Object Tracking VOT2021 Challenge Results},
	author={Kristan, Matej and Matas, Ji{\v{r}}{\'\i} and Leonardis, Ale{\v{s}} and Felsberg, Michael and Pflugfelder, Roman and K{\"a}m{\"a}r{\"a}inen, Joni-Kristian and Chang, Hyung Jin and Danelljan, Martin and Cehovin, Luka and Luke{\v{z}}i{\v{c}}, Alan and others},
	booktitle={ICCVW},
	pages={2711--2738},
	year={2021}
}

@article{liuchang1,
  title={Spatial-temporal initialization dilemma: towards realistic visual tracking},
  author={Liu, Chang and Yuan, Yongsheng and Chen, Xin and Lu, Huchuan and Wang, Dong},
  journal={Visual Intelligence},
  pages={35},
  year={2024}
}

@article{liuchang2,
  title={Long-term visual tracking: review and experimental comparison},
  author={Liu, Chang and Chen, Xiao-Fan and Bo, Chun-Juan and Wang, Dong},
  journal={Machine Intelligence Research},
  pages={512--530},
  year={2022}
}

@article{anti-uav,
  title={Vision-based anti-UAV detection and tracking},
  author={Zhao, Jie and Zhang, Jingshu and Li, Dongdong and Wang, Dong},
  journal={TITS},
  pages={25323--25334},
  year={2022}
}

@article{hcatm,
  author       = {Xin Chen and
                  Ben Kang and
                  Jiawen Zhu and
                  Dongdong Li and
                  Chunjuan Bo and
                  Dong Wang},
  title        = {Exploring a Hierarchical Cross-Attention Transformer for High-Speed
                  Tracking},
  journal      = {CVM},
  pages        = {1113--1132},
  year         = {2025}
}

@article{cmdtrack,
  title={Cross-Modality Distillation for Multi-Modal Tracking},
  author={Tianlu Zhang and
                  Qiang Zhang and
                  Kurt Debattista and
                  Jungong Han},
  journal=TPAMI,
  pages={5847--5865},
  year={2025}
}

@inproceedings{artrackv2,
  title={{ART}rack{V}2: Prompting Autoregressive Tracker Where to Look and How to Describe},
  author={Bai, Yifan and Zhao, Zeyang and Gong, Yihong and Wei, Xing},
  booktitle=CVPR,
  pages={19048--19057},
  year={2024}
}

@inproceedings{artrack,
  title={Autoregressive Visual Tracking},
  author={Wei, Xing and Bai, Yifan and Zheng, Yongchao and Shi, Dahu and Gong, Yihong},
  booktitle=CVPR,
  pages={9697--9706},
  year={2023}
}

@inproceedings{asymtrack,
  title={Two-Stream Beats One-Stream: Asymmetric Siamese Network for Efficient Visual Tracking},
  author={Zhu, Jiawen and Tang, Huayi and Chen, Xin and Wang, Xinying and Wang, Dong and Lu, Huchuan},
  booktitle=AAAI,
  pages={10959--10967},
  year={2025}
}

@inproceedings{uvltrack,
  title={Unifying Visual and Vision-Language Tracking via Contrastive Learning},
  author={Ma, Yinchao and Tang, Yuyang and Yang, Wenfei and Zhang, Tianzhu and Zhang, Jinpeng and Kang, Mengxue},
  booktitle=AAAI,
  pages={4107--4116},
  year={2024}
}

@inproceedings{TATrack-T,
  author       = {Hongyu Wang and
                  Xiaotao Liu and
                  Yifan Li and
                  Meng Sun and
                  Dian Yuan and
                  Jing Liu},
  title        = {Temporal Adaptive {RGBT} Tracking with Modality Prompt},
  booktitle    = AAAI,
  pages        = {5436--5444},
  year         = {2024}
}

@inproceedings{bat,
  author       = {Bing Cao and
                  Junliang Guo and
                  Pengfei Zhu and
                  Qinghua Hu},
  title        = {Bi-directional Adapter for Multi-modal Tracking},
  booktitle=AAAI,
  pages={927--935},
  year={2024}
}

@inproceedings{odtrack,
  title={O{D}track: Online Dense Temporal Token Learning for Visual Tracking},
  author={Zheng, Yaozong and Zhong, Bineng and Liang, Qihua and Mo, Zhiyi and Zhang, Shengping and Li, Xianxian},
  booktitle=AAAI,
  pages={7588--7596},
  year={2024}
}

@inproceedings{TATrack,
  title={Target-Aware Tracking with Long-Term Context Attention},
  author={He, Kaijie and Zhang, Canlong and Xie, Sheng and Li, Zhixin and Wang, Zhiwen},
  booktitle=AAAI,
  pages={773--780},
  year={2023}
}

@inproceedings{tbsi,
  title={Bridging Search Region Interaction with Template for {RGB-T} Tracking},
  author={Tianrui Hui and
                  Zizheng Xun and
                  Fengguang Peng and
                  Junshi Huang and
                  Xiaoming Wei and
                  Xiaolin Wei and
                  Jiao Dai and
                  Jizhong Han and
                  Si Liu},
  booktitle=CVPR,
  pages={13630--13639},
  year={2023}
}

@inproceedings{LoRAT,
  title={Tracking Meets LoRA: Faster Training, Larger Model, Stronger Performance},
  author={Lin, Liting and Fan, Heng and Zhang, Zhipeng and Wang, Yaowei and Xu, Yong and Ling, Haibin},
  booktitle=ECCV,
  pages={300-318},
  year={2024}
}

@article{fastitpn,
  title={Fast-iTPN: Integrally Pre-trained Transformer Pyramid Network with Token Migration},
  author={Tian, Yunjie and Xie, Lingxi and Qiu, Jihao and Jiao, Jianbin and Wang, Yaowei and Tian, Qi and Ye, Qixiang},
  journal=TPAMI,
    pages={1-15},
  year={2024}
}

@inproceedings{moe,
  author       = {Dmitry Lepikhin and
                  HyoukJoong Lee and
                  Yuanzhong Xu and
                  Dehao Chen and
                  Orhan Firat and
                  Yanping Huang and
                  Maxim Krikun and
                  Noam Shazeer and
                  Zhifeng Chen},
  title        = {GShard: Scaling Giant Models with Conditional Computation and Automatic
                  Sharding},
  booktitle    = ICLR,
  year         = {2021}
}

@inproceedings{aqatrack,
  title={Autoregressive Queries for Adaptive Tracking with Spatio-Temporal Transformers},
  author={Xie, Jinxia and Zhong, Bineng and Mo, Zhiyi and Zhang, Shengping and Shi, Liangtao and Song, Shuxiang and Ji, Rongrong},
  booktitle={CVPR},
  pages={19300--19309},
  year={2024}
}

@inproceedings{evptrack,
  title={Explicit Visual Prompts for Visual Object Tracking},
  author={Shi, Liangtao and Zhong, Bineng and Liang, Qihua and Li, Ning and Zhang, Shengping and Li, Xianxian},
  booktitle={AAAI},
  pages={4838--4846},
  year={2024}
}

@inproceedings{vasttrack,
  title={VastTrack: Vast Category Visual Object Tracking},
  author={Peng, Liang and Gao, Junyuan and Liu, Xinran and Li, Weihong and Dong, Shaohua and Zhang, Zhipeng and Fan, Heng and Zhang, Libo},
  booktitle=NIPS,
  pages={130797-130818},
  year={2024}
}

@inproceedings{cornernet,
  title={Corner{N}et: Detecting Objects as Paired Keypoints},
  author={Law, Hei and Deng, Jia},
  booktitle=ECCV,
  pages={734--750},
  year={2018}
}

@inproceedings{CiteTrack,
  title={Cite{T}racker: Correlating Image and Text for Visual Tracking},
  author={Li, Xin and Huang, Yuqing and He, Zhenyu and Wang, Yaowei and Lu, Huchuan and Yang, Ming-Hsuan},
  booktitle=ICCV,
  pages={9974--9983},
  year={2023}
}

@inproceedings{ostrack,
  title={Joint Feature Learning and Relation Modeling for Tracking: A One-Stream Framework},
  author={Ye, Botao and Chang, Hong and Ma, Bingpeng and Shan, Shiguang and Chen, Xilin},
  booktitle=ECCV,
  pages={341--357},
  year={2022},
}

@inproceedings{AiATrack,
  title={{AiATrack}: Attention in Attention for Transformer Visual Tracking},
  author={Gao, Shenyuan and Zhou, Chunluan and Ma, Chao and Wang, Xinggang and Yuan, Junsong},
  booktitle=ECCV,
  pages={146--164},
  year={2022}
}

@inproceedings{simtrack,
  title={Backbone is All Your Need: A Simplified Architecture for Visual Object Tracking},
  author={Chen, Boyu and Li, Peixia and Bai, Lei and Qiao, Lei and Shen, Qiuhong and Li, Bo and Gan, Weihao and Wu, Wei and Ouyang, Wanli},
  booktitle=ECCV,
  pages={375--392},
  year={2022}
}

@inproceedings{ToMP,
  title={Transforming Model Prediction for Tracking},
  author={Mayer, Christoph and Danelljan, Martin and Bhat, Goutam and Paul, Matthieu and Paudel, Danda Pani and Yu, Fisher and Van Gool, Luc},
  booktitle=CVPR,
  pages={8731--8740},
  year={2022}
}

@inproceedings{mixformer,
  title={MixFormer: End-to-End Tracking with Iterative Mixed Attention},
  author={Cui, Yutao and Jiang, Cheng and Wang, Limin and Wu, Gangshan},
  booktitle=CVPR,
  pages={13608--13618},
  year={2022}
}

@article{mixformer_journal,
  title={MixFormer: End-to-End Tracking with Iterative Mixed Attention},
  author={Cui, Yutao and Jiang, Cheng and Wang, Limin and Wu, Gangshan},
  journal=TPAMI,
  pages={0--18},
  year={2024}
}

@inproceedings{mixformerv2,
 author = {Cui, Yutao and Song, Tianhui and Wu, Gangshan and Wang, Limin},
 booktitle =NIPS,
 pages = {58736--58751},
 title = {MixFormerV2: Efficient Fully Transformer Tracking},
 year = {2023}
}

@inproceedings{HiT,
  title={Exploring Lightweight Hierarchical Vision Transformers for Efficient Visual Tracking},
  author={Kang, Ben and Chen, Xin and Wang, Dong and Peng, Houwen and Lu, Huchuan},
  booktitle=ICCV,
  pages={9612--9621},
  year={2023}
}

@inproceedings{fear,
  title={{FEAR}: {Fast, Efficient, Accurate and Robust Visual Tracker}},
  author={Borsuk, Vasyl and Vei, Roman and Kupyn, Orest and Martyniuk, Tetiana and Krashenyi, Igor and Matas, Ji{\v{r}}i},
  booktitle=ECCV,
  pages={644--663},
  year={2022}
}

@inproceedings{HCAT,
  title={Efficient {Visual Tracking via Hierarchical Cross-Attention Transformer}},
  author={Chen, Xin and Kang, Ben and Wang, Dong and Li, Dongdong and Lu, Huchuan},
  booktitle={ECCVW},
  pages={461--477},
  year={2022}
}

@inproceedings{ETTrack,
  title={Efficient {Visual Tracking with Exemplar Transformers}},
  author={Blatter, Philippe and Kanakis, Menelaos and Danelljan, Martin and Van Gool, Luc},
  booktitle={WACV},
  pages={1571--1581},
  year={2023}
}

@inproceedings{lighttrack,
  title={{LightTrack: Finding Lightweight Neural Networks for Object Tracking
  via One-Shot Architecture Search}},
  author={Yan, Bin and Peng, Houwen and Wu, Kan and Wang, Dong and Fu, Jianlong and Lu, Huchuan},
  booktitle=CVPR,
  pages={15180--15189},
  year={2021}
}

@inproceedings{vmoe,
  title={Scaling Vision with Sparse Mixture of Experts},
  author={Riquelme, Carlos and Puigcerver, Joan and Mustafa, Basil and Neumann, Maxim and Jenatton, Rodolphe and Susano Pinto, Andr{\'e} and Keysers, Daniel and Houlsby, Neil},
  booktitle=NIPS,
  pages={8583--8595},
  year={2021}
}

@inproceedings{swintransformer,
  title={Swin Transformer: Hierarchical Vision Transformer Using Shifted Windows},
  author={Liu, Ze and Lin, Yutong and Cao, Yue and Hu, Han and Wei, Yixuan and Zhang, Zheng and Lin, Stephen and Guo, Baining},
  booktitle=ICCV,
  pages={10012--10022},
  year={2021}
}

@inproceedings{Stark,
  title={Learning Spatio-Temporal Transformer for Visual Tracking},
  author={Yan, Bin and Peng, Houwen and Fu, Jianlong and Wang, Dong and Lu, Huchuan},
  booktitle=ICCV,
  pages={10448--10457},
  year={2021}
}

@inproceedings{MDNet,
  title={Learning Multi-domain Convolutional Neural Networks for Visual Tracking},
  author={Nam, Hyeonseob and Han, Bohyung},
  booktitle=CVPR,
  pages={4293--4302},
  year={2016}
}

@inproceedings{PrDiMP,
  title={Probabilistic Regression for Visual Tracking},
  author={Danelljan, Martin and Gool, Luc Van and Timofte, Radu},
  booktitle=CVPR,
  pages={7183--7192},
  year={2020}
}

@inproceedings{AdamW,
  title={Decoupled Weight Decay Regularization},
  author={Loshchilov, Ilya and Hutter, Frank},
  booktitle=ICLR,
  pages={1--9},
  year={2018},
}

@article{dyhit,
  title={Exploiting Lightweight Hierarchical ViT and Dynamic Framework for Efficient Visual Tracking},
  author={Kang, Ben and Chen, Xin and Zhao, Jie and Bo, Chunjuan and Wang, Dong and Lu, Huchuan},
  journal=IJCV,
  pages={1--23},
  year={2025},
}

@inproceedings{tctrack,
  title={TCTrack: Temporal Contexts for Aerial Tracking},
  author={Cao, Ziang and Huang, Ziyuan and Pan, Liang and Zhang, Shiwei and Liu, Ziwei and Fu, Changhong},
  booktitle={CVPR},
  pages={14778-14788},
  year={2022}
}

@inproceedings{COCO,
  title={{Microsoft COCO}: Common Objects in Context},
  author={Tsung-Yi Lin and Michael Maire and Serge J. Belongie and Lubomir D. Bourdev and Ross B. Girshick and James Hays and Pietro Perona and Deva Ramanan and Piotr Doll{\'a}r and C. Lawrence Zitnick},
  booktitle=ECCV,
  pages={740--755},
  year={2014},
}

@inproceedings{ATOM,
  author    = {Martin Danelljan and Goutam Bhat and Fahad Shahbaz Khan and Michael Felsberg},
  title     = {{ATOM: Accurate} Tracking by Overlap Maximization},
  booktitle = CVPR,
  pages={4660--4669},
  year={2019}
}

@article{spot,
  title={Spot-Adaptive Knowledge Distillation},
  author={Song, Jie and Chen, Ying and Ye, Jingwen and Song, Mingli},
  journal=TIP,
  pages={3359--3370},
  year={2022}
}

@inproceedings{DiMP,
  author    = {Goutam Bhat and Martin Danelljan and Luc Van Gool and Radu Timofte},
  title     = {Learning Discriminative Model Prediction for Tracking},
  booktitle = ICCV,
  pages={6182--6191},
  year={2019}
}

@article{lasot_journal,
  title={{LaSOT}: A High-Quality Large-Scale Single Object Tracking Benchmark},
  author={Fan, Heng and Bai, Hexin and Lin, Liting and Yang, Fan and Chu, Peng and Deng, Ge and Yu, Sijia and Huang, Mingzhen and Liu, Juehuan and Xu, Yong and others},
  journal=IJCV,
  pages={439--461},
  year={2021}
}

@InProceedings{LaSOT,
author = {Heng Fan and Liting Lin and Fan Yang and Peng Chu and Ge Deng and Sijia Yu and Hexin Bai and Yong Xu and Chunyuan Liao and Haibin Ling},
title = {{LaSOT}: A High-Quality Benchmark for Large-Scale Single Object Tracking},
booktitle = CVPR,
  pages={5374--5383},
  year={2019}
}

@InProceedings{ResNet,
author = {Kaiming He and Xiangyu Zhang and Shaoqing Ren and Jian Sun},
title = {Deep Residual Learning for Image Recognition},
booktitle = CVPR,
  pages={770--778},
  year={2016}
}

@inproceedings{softmoe,
  author       = {Joan Puigcerver and
                  Carlos Riquelme Ruiz and
                  Basil Mustafa and
                  Neil Houlsby},
  title        = {From Sparse to Soft Mixtures of Experts},
  booktitle    = {ICLR},
  year         = {2024},
}

@article{emoe,
  title={eMoE-Tracker: Environmental MoE-Based Transformer for Robust Event-Guided
                  Object Tracking},
  author={Chen, Yucheng and Wang, Lin},
  journal={IEEE RAL},
  pages= {1393--1400},
  year={2024},
}

@article{moetracker,
  title={Revisiting {RGBT} Tracking Benchmarks from the Perspective of Modality
                  Validity: {A} New Benchmark, Problem, and Solution},
  author={Tang, Zhangyong and Xu, Tianyang and Feng, Zhen-Hua and Zhu, Xuefeng and Wang, He and Shao, Pengcheng and Cheng, Chunyang and Wu, Xiaojun and Awais, Muhammad and Atito, Sara and others},
  journal={CoRR},
  year={2024}
}

@article{GOT10K,
  title={{GOT}-10k: A Large High-Diversity Benchmark for Generic Object Tracking in the Wild},
  author={Huang, Lianghua and Zhao, Xin and Huang, Kaiqi},
  journal=TPAMI,
  pages={1562--1577},
  year={2019},
}

@inproceedings{SiamBAN,
  title={Siamese Box Adaptive Network for Visual Tracking},
  author={Chen, Zedu and Zhong, Bineng and Li, Guorong and Zhang, Shengping and Ji, Rongrong},
  booktitle=CVPR,
  pages={6668--6677},
  year={2020}
}

@inproceedings{mcitrack,
  title={Exploring Enhanced Contextual Information for Video-Level Object Tracking},
  author={Kang, Ben and Chen, Xin and Lai, Simiao and Liu, Yang and Liu, Yi and Wang, Dong},
  booktitle=AAAI,
  pages={4194--4202},
  year={2025}
}

@inproceedings{sutrack,
  title={Sutrack: Towards Simple and Unified Single Object Tracking},
  author={Chen, Xin and Kang, Ben and Geng, Wanting and Zhu, Jiawen and Liu, Yi and Wang, Dong and Lu, Huchuan},
  booktitle=AAAI,
  pages={2239--2247},
  year={2025}
}

@inproceedings{AlexNet,
  title={Imagenet Classification with Deep Convolutional Neural Networks},
  author={Krizhevsky, Alex and Sutskever, Ilya and Hinton, Geoffrey E},
  booktitle=NIPS,
  pages={1106-1114},
  year={2012}
}

@inproceedings{trackingnet,
  title={Tracking{N}et: A Large-Scale Dataset and Benchmark for Object Tracking in the Wild},
  author={Muller, Matthias and Bibi, Adel and Giancola, Silvio and Alsubaihi, Salman and Ghanem, Bernard},
  booktitle=ECCV,
  pages={300--317},
  year={2018}
}

@inproceedings{clip,
  title={Learning Transferable Visual Models from Natural Language Supervision},
  author={Radford, Alec and Kim, Jong Wook and Hallacy, Chris and Ramesh, Aditya and Goh, Gabriel and Agarwal, Sandhini and Sastry, Girish and Askell, Amanda and Mishkin, Pamela and Clark, Jack and others},
  booktitle={ICML},
  pages={8748--8763},
  year={2021},
}

@inproceedings{gumbelsoftmax,
  author       = {Eric Jang and
                  Shixiang Gu and
                  Ben Poole},
  title        = {Categorical Reparameterization with Gumbel-Softmax},
  booktitle    = {ICLR},
  year         = {2017},
}

@inproceedings{spmtrack,
  title={SPMTrack: Spatio-Temporal Parameter-Efficient Fine-Tuning with Mixture of Experts for Scalable Visual Tracking},
  author={Cai, Wenrui and Liu, Qingjie and Wang, Yunhong},
  booktitle={CVPR},
  pages={16871--16881},
  year={2025}
}

@article{kd,
  title={Distilling the Knowledge in a Neural Network},
  author={Hinton, Geoffrey and Vinyals, Oriol and Dean, Jeff},
  journal={arXiv preprint arXiv:1503.02531},
  year={2015}
}

@article{distbert,
  title={DistilBERT, a distilled version of BERT: smaller, faster, cheaper and lighter},
  author={Sanh, Victor and Debut, Lysandre and Chaumond, Julien and Wolf, Thomas},
  journal={arXiv preprint arXiv:1910.01108},
  year={2019}
}

@inproceedings{at,
  author       = {Sergey Zagoruyko and
                  Nikos Komodakis},
  title        = {Paying More Attention to Attention: Improving the Performance of Convolutional
                  Neural Networks via Attention Transfer},
  booktitle    = {ICLR},
  year         = {2017},
}

@inproceedings{rkd,
  title={Relational Knowledge Distillation},
  author={Park, Wonpyo and Kim, Dongju and Lu, Yan and Cho, Minsu},
  booktitle={CVPR},
  pages={3967--3976},
  year={2019}
}

@inproceedings{skd,
  title={Similarity-Preserving Knowledge Distillation},
  author={Tung, Frederick and Mori, Greg},
  booktitle={ICCV},
  pages={1365--1374},
  year={2019}
}

@inproceedings{dkd,
  title={Decoupled Knowledge Distillation},
  author={Zhao, Borui and Cui, Quan and Song, Renjie and Qiu, Yiyu and Liang, Jiajun},
  booktitle={CVPR},
  pages={11953--11962},
  year={2022}
}

@inproceedings{ViT,
  title={An Image is Worth 16x16 Words: Transformers for Image Recognition at Scale},
  author={Dosovitskiy, Alexey and Beyer, Lucas and Kolesnikov, Alexander and Weissenborn, Dirk and Zhai, Xiaohua and Unterthiner, Thomas and Dehghani, Mostafa and Minderer, Matthias and Heigold, Georg and Gelly, Sylvain and others},
  booktitle=ICLR,
  year={2020}
}

@inproceedings{fitnet,
  author       = {Adriana Romero and
                  Nicolas Ballas and
                  Samira Ebrahimi Kahou and
                  Antoine Chassang and
                  Carlo Gatta and
                  Yoshua Bengio},
  title        = {FitNets: Hints for Thin Deep Nets},
  booktitle    = {ICLR},
  year         = {2015},
}

@inproceedings{GIoU,
  author    = {Hamid Rezatofighi and Nathan Tsoi and JunYoung Gwak and
               Amir Sadeghian and Ian D. Reid and Silvio Savarese},
  title     = {Generalized Intersection Over Union: {A} Metric and a Loss for Bounding Box Regression},
  booktitle = CVPR,
  pages={658--666},
  year={2019}
}

@inproceedings{vipt,
  title={Visual Prompt Multi-Modal Tracking},
  author={Zhu, Jiawen and Lai, Simiao and Chen, Xin and Wang, Dong and Lu, Huchuan},
  booktitle=CVPR,
  pages={9516--9526},
  year={2023}
}

@inproceedings{OneTracker,
    author    = {Hong, Lingyi and Yan, Shilin and Zhang, Renrui and Li, Wanyun and Zhou, Xinyu and Guo, Pinxue and Jiang, Kaixun and Chen, Yiting and Li, Jinglun and Chen, Zhaoyu and Zhang, Wenqiang},
    title     = {OneTracker: Unifying Visual Object Tracking with Foundation Models and Efficient Tuning},
    booktitle = CVPR,
    year      = {2024},
    pages     = {19079-19091}
}

@inproceedings{SDSTrack,
    author    = {Hou, Xiaojun and Xing, Jiazheng and Qian, Yijie and Guo, Yaowei and Xin, Shuo and Chen, Junhao and Tang, Kai and Wang, Mengmeng and Jiang, Zhengkai and Liu, Liang and Liu, Yong},
    title     = {SDSTrack: Self-Distillation Symmetric Adapter Learning for Multi-Modal Visual Object Tracking},
    booktitle = CVPR,
    year      = {2024},
    pages     = {26551-26561}
}

@InProceedings{untrack,
    author    = {Wu, Zongwei and Zheng, Jilai and Ren, Xiangxuan and Vasluianu, Florin-Alexandru and Ma, Chao and Paudel, Danda Pani and Van Gool, Luc and Timofte, Radu},
    title     = {Single-Model and Any-Modality for Video Object Tracking},
    booktitle = CVPR,
    year      = {2024},
    pages     = {19156-19166}
}

@article{seqtrackv2,
      title={Unified Sequence-to-Sequence Learning for Single- and Multi-Modal Visual Object Tracking}, 
      author={Xin Chen and Ben Kang and Jiawen Zhu and Dong Wang and Houwen Peng and Huchuan Lu},
      journal={arXiv preprint arXiv:2304.14394},
      year={2024}
}

@inproceedings{protrack,
	title={Prompting for Multi-Modal Tracking},
	author={Yang, Jinyu and Li, Zhe and Zheng, Feng and Leonardis, Ales and Song, Jingkuan},
	booktitle={ACMMM},
	pages={3492--3500},
	year={2022}
}

@inproceedings{apfnet,
	title={Attribute-based Progressive Fusion Network for {RGBT} Tracking},
	author={Xiao, Yun and Yang, Mengmeng and Li, Chenglong and Liu, Lei and Tang, Jin},
	booktitle = {AAAI},
	pages={2831--2838},
	year={2022}
}

@article{jmmac,
	title={Jointly Modeling Motion and Appearance Cues for Robust {RGB-T} Tracking},
	author={Zhang, Pengyu and Zhao, Jie and Bo, Chunjuan and Wang, Dong and Lu, Huchuan and Yang, Xiaoyun},
	journal=TIP,
	pages={3335--3347},
	year={2021},
	publisher={IEEE}
}

@inproceedings{rgbd1k,
	title={{RGBD1K}: A Large-Scale Dataset and Benchmark for {RGB-D} Object Tracking},
	author={Zhu, Xue-Feng and Xu, Tianyang and Tang, Zhangyong and Wu, Zucheng and Liu, Haodong and Yang, Xiao and Wu, Xiao-Jun and Kittler, Josef},
	booktitle={AAAI},
	pages={3870--3878},
	year={2023}
}

@inproceedings{dal,
	title={DAL: A Deep Depth-Aware Long-term Tracker},
	author={Qian, Yanlin and Yan, Song and Luke{\v{z}}i{\v{c}}, Alan and Kristan, Matej and K{\"a}m{\"a}r{\"a}inen, Joni-Kristian and Matas, Ji{\v{r}}{\'\i}},
	booktitle={ICPR},
	pages={7825--7832},
	year={2021},
}

@article{ca3dms,
	title={Context-Aware Three-Dimensional Mean-Shift with Occlusion Handling for Robust Object Tracking in {RGB-D} Videos},
	author={Liu, Ye and Jing, Xiao-Yuan and Nie, Jianhui and Gao, Hao and Liu, Jun and Jiang, Guo-Ping},
	journal=TMM,
	pages={664--677},
	year={2018},
	publisher={IEEE}
}

@inproceedings{JointNLT,
  title={Joint Visual Grounding and Tracking with Natural Language Specification},
  author={Zhou, Li and Zhou, Zikun and Mao, Kaige and He, Zhenyu},
  booktitle=CVPR,
  pages={23151--23160},
  year={2023}
}

@inproceedings{DecoupleTNL,
  title={Tracking by Natural Language Specification with Long Short-term Context Decoupling},
  author={Ma, Ding and Wu, Xiangqian},
  booktitle={ICCV},
  pages={14012--14021},
  year={2023}
}

@article{zhao2023transformervision-languagetracking,
  title={Transformer Vision-Language Tracking via Proxy Token Guided Cross-Modal Fusion},
  author={Zhao, Haojie and Wang, Xiao and Wang, Dong and Lu, Huchuan and Ruan, Xiang},
  journal={PRL},
  pages={10--16},
  year={2023}
}

@inproceedings{VLTTT,
  title={Divert More Attention to Vision-Language Tracking},
  author={Guo, Mingzhe and Zhang, Zhipeng and Fan, Heng and Jing, Liping},
  booktitle=NIPS,
  pages={4446--4460},
  year={2022}
}

@inproceedings{CapsuleTNL,
  title={Capsule-based Object Tracking with Natural Language Specification},
  author={Ma, Ding and Wu, Xiangqian},
  booktitle=ACMMM,
  pages={1948--1956},
  year={2021}
}

@inproceedings{TNLS,
  title={Tracking by Natural Language Specification},
  author={Li, Zhenyang and Tao, Ran and Gavves, Efstratios and Snoek, Cees GM and Smeulders, Arnold WM},
  booktitle={CVPR},
  pages={6495--6503},
  year={2017}
}

@inproceedings{SNLT,
  title={Siamese Natural Language Tracker: Tracking by Natural Language Descriptions with Siamese Trackers},
  author={Feng, Qi and Ablavsky, Vitaly and Bai, Qinxun and Sclaroff, Stan},
  booktitle={CVPR},
  pages={5851--5860},
  year={2021}
}

@article{emtrack,
  title={Emtrack: Efficient Multimodal Object Tracking},
  author={Liu, Chang and Guan, Ziqi and Lai, Simiao and Liu, Yang and Lu, Huchuan and Wang, Dong},
  journal=TCSVT,
  pages={2202--2214},
  year={2024},
  publisher={IEEE}
}

@inproceedings{RTTNLD,
  title={Real-Time Visual Object Tracking with Natural Language Description},
  author={Feng, Qi and Ablavsky, Vitaly and Bai, Qinxun and Li, Guorong and Sclaroff, Stan},
  booktitle={WACV},
  pages={700--709},
  year={2020}
}

@inproceedings{smat,
  title={Separable Self and Mixed Attention Transformers for Efficient Object Tracking},
  author={Gopal, Goutam Yelluru and Amer, Maria A},
  booktitle={WACV},
  pages={6708--6717},
  year={2024}
}

@article{rgbt234,
	title={{RGB-T} Object Tracking: Benchmark and Baseline},
	author={Li, Chenglong and Liang, Xinyan and Lu, Yijuan and Zhao, Nan and Tang, Jin},
	journal=PR,
	pages={106977},
	year={2019},
	publisher={Elsevier}
}

@article{lasher,
	title={Las{H}e{R}: A Large-Scale High-Diversity Benchmark for {RGBT} Tracking},
	author={Li, Chenglong and Xue, Wanlin and Jia, Yaqing and Qu, Zhichen and Luo, Bin and Tang, Jin and Sun, Dengdi},
	journal=TIP,
	pages={392--404},
	year={2021},
	publisher={IEEE}
}

@article{visevent,
  author={Wang, Xiao and Li, Jianing and Zhu, Lin and Zhang, Zhipeng and Chen, Zhe and Li, Xin and Wang, Yaowei and Tian, Yonghong and Wu, Feng},
  journal=TCYB, 
  title={VisEvent: Reliable Object Tracking via Collaboration of Frame and Event Flows}, 
  year={2024},
  pages={1997-2010}}

@inproceedings{depthtrack,
	title={Depth{T}rack: Unveiling the Power of {RGBD} Tracking},
	author={Yan, Song and Yang, Jinyu and K{\"a}pyl{\"a}, Jani and Zheng, Feng and Leonardis, Ale{\v{s}} and K{\"a}m{\"a}r{\"a}inen, Joni-Kristian},
	booktitle={ICCV},
	pages={10725--10733},
	year={2021}
}

@inproceedings{hift,
  title={Hift: Hierarchical Feature Transformer for Aerial Tracking},
  author={Cao, Ziang and Fu, Changhong and Ye, Junjie and Li, Bowen and Li, Yiming},
  booktitle={ICCV},
  pages={15457--15466},
  year={2021}
}

@inproceedings{litetrack,
  title={LiteTrack: Layer Pruning with Asynchronous Feature Extraction for Lightweight and Efficient Visual Tracking},
  author={Wei, Qingmao and Zeng, Bi and Liu, Jianqi and He, Li and Zeng, Guotian},
  booktitle={ICRA},
  pages={4968--4975},
  year={2024},
}
}


\end{document}